\documentclass{article} 
\usepackage{iclr2027_conference,times}

\PassOptionsToPackage{hyphens}{url}
\usepackage{hyperref}
\usepackage{url}

\usepackage{graphicx}
\usepackage{booktabs}
\usepackage{amsmath}
\usepackage{amssymb}
\usepackage{placeins} 
\title{ROAR: Unifying Runs across Heterogeneous \\ AI-Driven Research Systems}

\iclrfinalcopy
\author{Leo Y. Lin\thanks{Work done while at IBM Research.} \\
Purdue University \\
\texttt{lin1736@purdue.edu} \\
\And
Vishakha Ramani \\
IBM Research \\
\texttt{vishakharamani@ibm.com} \\
\And
Z. Berkay Celik \\
Purdue University \\
\texttt{zcelik@purdue.edu} \\
\And
Paul Castro \\
IBM Research \\
\texttt{castrop@us.ibm.com} \\
\And
Marquita Ellis \\
IBM Research \\
\texttt{mme@berkeley.edu} \\
}

\begin{document}

\maketitle

\begin{abstract}
Each run of an AI-driven research system (ADRS) is an expensive search over a vast solution space, and dependable evaluation requires many runs, making run data both costly to produce and valuable to retain for large-scale analysis. Yet this data remains fragmented: teams operate in isolation, ADRS frameworks emit results in different formats, and no shared infrastructure exists to aggregate or compare runs across problems and systems. We present ROAR, a solution for systematically unifying and analyzing heterogeneous ADRS outputs. ROAR addresses two challenges: reconciling heterogeneous ADRS outputs and enabling analytics across runs with different objectives and scoring functions. We achieve this through a relational schema and parsing layer that normalize heterogeneous ADRS outputs while preserving data lineage and temporal structure, and accommodating new systems without requiring schema modifications. From building a corpus of more than 900 runs from multiple ADRS, we show how pooled data can reveal properties of problem landscapes that are difficult to observe. Consistent with prior work, runs with identical configurations may converge to different scores. We find that many runs realize most gains early, and that the effectiveness of different strategies for incorporating prior solutions into the search process varies across problems. We further show that the pooled corpus is actionable and not merely analytical by using ROAR to configure ADRS runs. Together, these results illustrate how pooled ADRS data can expose problem-dependent structure in search behavior that is difficult to detect from any single system, team, or benchmark.  Such cross-cutting insights are difficult to obtain while runs remain siloed; ROAR is the first infrastructure designed to unify them.
\end{abstract}

\section{Introduction}\label{sec:intro}

AI-Driven Research Systems (ADRS) have emerged as a powerful paradigm for automated problem solving, combining large language model generators with iterative optimization mechanisms to discover novel solutions across domains ranging from database transaction optimization to model placement~\citep{cheng_barbarians_2025, cheng_let_2025}.
As adoption grows, practitioners independently generate vast amounts of ADRS run data spanning diverse problems and configurations.

Pooling these data could yield immediate practical benefits by identifying promising and ineffective configurations before costly experiments are repeated, thereby reducing both time and computational expense. More importantly, patterns across aggregated runs may enable deeper insights into ADRS behavior, revealing the principles that determine when optimization succeeds, stagnates, or becomes trapped in suboptimal regions. Such insights could support breakthroughs in the design, analysis, and application of more effective ADRS systems.
Despite these, a significant gap remains: there is no systematic way to aggregate, share, and analyze cross-domain ADRS data.

Addressing this gap is challenging for two reasons. First, ADRS frameworks, such as SkyDiscover~\citep{liu_skydiscover_2026}, OpenEvolve~\citep{openevolve}, and ShinkaEvolve~\citep{lange_shinkaevolve_2025}, produce outputs using different data formats and semantic abstractions. Extracting meaningful, comparable data at scale from these heterogeneous outputs requires careful abstraction. 
Second, even with a unified representation, it is unclear how to perform useful analytics when the assessors and problems differ across runs.

We present ROAR (Repository of ADRS Runs)\footnote{https://github.com/AI-native-Systems-Research/ROAR}, a unified data infrastructure for aggregating, storing, and analyzing outputs from multiple ADRS to address the challenges.
First, we extend prior work on abstracting ADRS components and dynamics~\citep{ellis_dont_2026} to introduce a unified relational database schema for representing runs across heterogeneous frameworks. We complement this schema with a parser layer that translates each framework’s native outputs into a common representation. This architecture enables cross-system aggregation, preserves lineage and temporal structure, and supports new systems without schema changes. The relational design is also agent-friendly: its explicit, introspectable structure supports reliable query generation, while augmenting with \textit{pgvector} enables vector-based operations. 
Second, we demonstrate the utility of such abstracted data storage by introducing a three-tier analytical paradigm that derives insights even when evaluators and problems differ across runs. Furthermore, we present a data-driven procedure that estimates budget-allocation policies from pooled runs to configure ADRS runs.

We use ROAR to investigate four research questions about ADRS behavior:

\noindent\textbf{RQ1: Basin Structure.} Do identical configurations produce multimodal final score distributions, suggesting multiple attractors in the optimization landscape?

\noindent\textbf{RQ2: Stagnation Heuristics.} How far do ADRS escape rates fall below what the classical rule-of-3 implies, given that history-driven context growth violates its i.i.d.\ assumption?

\noindent\textbf{RQ3: Diversity and Context Composition.} Do the diversity of a run's early candidates and the composition of the context shown to the generator predict final scores and improvements?

\noindent\textbf{RQ4: Budget Allocation.} Can pooled runs tell a practitioner how to divide a fixed iteration budget so that it measurably outperforms current practice at equal compute?

Our main contributions are:
\begin{itemize}
    \item The first general data aggregation infrastructure for ADRS: a unified relational schema with a parser layer, enabling the cross-problem and cross-system analysis that RQ1--RQ4 require.
    \item A dataset with over 900 runs spanning 7 problems, 14 generator models, and 6 optimization mechanisms across multiple ADRS frameworks, including OpenEvolve, SkyDiscover, and GEPA, contributed by independent researchers.
    \item Exploratory data analysis demonstrating the infrastructure's utility by revealing problem landscape characteristics (multimodal basin structures, RQ1), ADRS design and utility insights (stopping heuristics, RQ2; context composition strategies, RQ3), and a data-driven budget allocation policy (RQ4).
\end{itemize}

\section{Background and Related Work}
\label{sec:background}

AI-Driven Research Systems (ADRS) place a language model inside an iterative search loop with a systematic evaluator, from early systems such as FunSearch~\citep{Romera-Paredes2024} and AlphaEvolve~\citep{novikov_alphaevolve_2025} to open source frameworks such as OpenEvolve~\citep{openevolve}, GEPA~\citep{agrawal_gepa_2026}, and SkyDiscover~\citep{liu_skydiscover_2026}. Prior analyses of ADRS behavior~\citep{cheng_barbarians_2025,cheng_let_2025,ellis_dont_2026,gupta2026} rely primarily on runs generated by the authors, while each framework records run history in different formats that generic experiment trackers such as MLflow~\citep{zaharia_mlflow_2018} do not reconcile. Appendix~\ref{sec:appendix-related-work} provides an extended discussion.

\section{Method}
\label{sec:method}

ROAR has four components: (1) a database schema that abstracts over heterogeneous system outputs while preserving analytical utility, (2) a repository that ingests raw outputs into that schema through a parser layer, including outputs contributed by coding agents, (3) scalable analytics pipelines that enable meaningful analysis even when graders and problems differ across runs, and (4) a procedure that leverages the pooled corpus to derive a run configuration. Components (1)--(3) are what make RQ1--RQ4 answerable across frameworks; component (4) is the procedure evaluated in RQ4.

\subsection{Database Schema Design}

A central challenge in studying AI-Driven Research System behavior across systems is that each framework produces outputs in different formats with different semantics. Whether the system uses population-based evolution with parent/context lineage (e.g., SkyDiscover, OpenEvolve) or sequential hill-climbing with iteration-by-iteration metrics (e.g., GEPA), the raw outputs vary in structure and naming conventions.

Despite these surface differences, all ADRS share the common conceptual structure formalized by the GAMBLe framework~\citep{ellis_dont_2026}: a \textit{Generator} $G$ that proposes candidate solutions, an \textit{Assessor} $\mathcal{A}$ that evaluates them, a \textit{Mechanism} $\mathcal{M}$ that governs selection and iteration, and a \textit{Budget} $B$ constraining computational resources. This decomposition motivates our schema design: systems execute \textit{campaigns} (a specific $(G, \mathcal{A}, \mathcal{M}, B)$ configuration applied to a problem) that produce \textit{candidates} (outputs from $G$) which receive \textit{measurements} (scores from $\mathcal{A}$).

We designed a six-table relational schema that captures this common structure while preserving system-specific details. The schema comprises of the following: \textbf{Systems}, recording the ADRS framework and version; \textbf{Campaigns}, storing run-level metadata (problem, algorithm, generator model, timestamps, and configuration); \textbf{Candidates}, the central abstraction representing generated solutions indexed by iteration order within a campaign; \textbf{Candidate Edges}, capturing lineage via typed edges distinguishing \emph{parent} links (direct mutation ancestry) from \emph{context} links (candidates shown as examples during generation); \textbf{Measurements}, storing assessor outputs as flexible key-value pairs, accommodating heterogeneous metrics without schema changes; and \textbf{Artifacts}, referencing raw files (source code, configurations) on disk, preserving full reproducibility without bloating the database. The full schema specification appears in Appendix~\ref{sec:appendix_schema}.

Our schema is designed to provide a consistent representation across ADRS frameworks, support reliable cross system analysis, and facilitate the use of coding agents for tasks such as data exploration, query generation, and automated analysis. To support these goals, we use a relational PostgreSQL schema rather than a document store for three reasons. First, the schema enforces structural consistency at write time, ensuring that data is normalized before entering the system rather than requiring defensive handling in downstream queries. Second, relational databases expose tables, column types, constraints, and relationships through schema introspection, allowing agents to inspect the database structure directly before constructing queries. In contrast, document stores often require agents to sample documents and infer their structure. Third, prior work suggests that LLMs generate SQL queries more accurately than document store query languages~\citep{sm3texttoquery}.

To retain the flexibility typically associated with document stores, we augment the relational core with three capabilities.
Vectors stored as a column directly via the pgvector extension, enabling in-database similarity computations alongside relational joins. 
Flexible key-value measurements that accommodate new metrics without schema migrations. References to raw artifacts on disk, preserving full fidelity for analyses requiring original data.

The schema supports a parser-based ingestion architecture in which each system has a dedicated parser that converts raw ADRS outputs into a common Pydantic intermediate representation before database insertion. 
This separation allows new systems with different output structures to be added without modifying the database layer, and provides type-safe validation at parse time.

\subsection {AI-Driven Evolving Repository}

The repository exposes an API for data ingestion that supports both traditional uploads and agent-driven contributions. Upload instructions and schema references are served as structured, machine-readable endpoints designed to be consumed directly by coding agents. A coding agent reads the workflow instructions, inspects its local campaign data, and selects the appropriate ingestion path.

For frameworks with known output structures, the system provides deterministic parsers that map raw files to the common schema. For unrecognized formats, the agent retrieves the target schema reference, explores the source data, and writes a one-off parser that maps the campaign outputs to the repository's intermediate representation. The resulting structured file is uploaded alongside the raw data, and both paths validate through the same Pydantic schema layer before insertion. This design reflects the broader shift toward agent-mediated workflows: the same coding agents that execute ADRS systems can also contribute their outputs to a shared repository with minimal human oversight. Embedding computation and system throughput are reported in Appendix~\ref{app:performance}.

\subsection{Analytical Pipeline}

A central analytical challenge is that different ADRS runs employ different assessors $\mathcal{A}$ producing scores on incomparable scales, precluding na\"ive pooling of raw values. We address this by introducing a three-tiered analytical paradigm that leverages progressively weaker assumptions about score comparability. At the first tier, we explore absolute comparability: when runs share the same assessor and problem, scores are directly comparable across campaigns, enabling distributional analyses such as detecting whether a configuration produces multimodal outcomes (RQ1). At the second tier, we leverage ordinal comparability: even when assessors differ, the event ``did the score improve?'' depends only on the ordinal structure of a single assessor, not on cross-run alignment of magnitudes. For example, how quickly runs escape stagnation does not require a shared scoring scale (RQ2). At the third tier, we propose assessor-agnostic metrics derived from stored artifacts rather than scores to bypass the assessor entirely, such as quantifying algorithmic diversity via embedding similarity on code artifacts (RQ3).

\subsection{Using ROAR for Budget Allocation}\label{sec:budget-allocation}

\textbf{The policy.} A practitioner supplies a problem, a generator, a mechanism, and an iteration budget $B$. ROAR returns a budget-allocation policy $\pi = (k_{\text{stop}}, n)$ according to stored campaigns of the same configuration, executed as follows: cap each campaign at $\lfloor B/n \rfloor$ iterations; abandon a campaign once its best score has stood unchanged for $k_{\text{stop}}$ consecutive iterations, spend the freed budget on a fresh campaign, and report the best score any campaign achieved. To judge how good a policy is, we average this best score over repeated executions of the policy,
\begin{equation}
V(\pi) = \mathbb{E}\biggl[\,\max_{c \in \mathcal{C}(\pi)} s^{(c)}_{\tau_c}\biggr],
\label{eq:policy-value}
\end{equation}
where $\mathcal{C}(\pi)$ is the set of campaigns launched before the budget runs out, $\tau_c$ is the iteration at which campaign $c$ is stopped, and $s^{(c)}_{\tau_c}$ is campaign $c$'s running best score at that point.
The baseline is a single campaign of $B$ iterations with no stopping. Fitting the policy optimizes $V(\pi)$ as written. In the evaluation, we write $c_1$ for the first campaign the policy launches and estimate $V(\pi \mid c_1 = c)$ for each stored campaign $c$, averaging over $c$ recovers $V(\pi)$.

\textbf{Fitting $\pi$ by trajectory replay.} Fitting requires two quantities for each stored campaign: the iteration at which the policy stops the campaign, and the running best score $s_t$ at that iteration. ROAR records the iteration and score of every candidate, so both quantities can be read directly from the stored campaigns. We select $\pi$ by grid search over 9 log-spaced values of $k_{\text{stop}}$ spanning $[2, B-1]$ and log-spaced values of $n$ spanning $[1, \lfloor B/10 \rfloor]$, and return $\pi$ with the highest $V(\pi)$. Since $V(\pi)$ is an expectation over executions, we estimate it for each candidate by replaying $T$ executions against stored campaigns and averaging the best score obtained. 

We measure the approximate cost of fitting in the number of score comparisons. Let $N$ be the number of stored campaigns of the input configuration. Computing running bests requires at most $B$ comparisons per campaign, or $NB$ in total. The grid holds at most $B$ candidate policies. Each candidate policy is executed $T$ times. A single execution launches at most $B$ campaigns and compares to find the best. Evaluating the grid thus requires at most $TB^2$ comparisons. Fitting therefore costs at most $NB + TB^2$ comparisons; in practice, this cost grows linearly in the number of pooled campaigns $N$. The generation and evaluation cost of every campaign was incurred when it was contributed to ROAR.

\section{Empirical Analysis}
\label{sec:experiments}

We evaluate the utility of ROAR through the four research questions posed in \S\ref{sec:intro}. RQ1--RQ3 are analyses of the pooled corpus spanning optimization dynamics and predictive analysis; RQ4 closes the loop by putting the corpus to use.

\subsection{Experimental Setup}

We aggregated outputs from multiple independent researchers using AI-Driven Research Systems for their problems into our database schema.
The current corpus holds 928 campaigns and 48,210 candidates; Appendix~\ref{sec:appendix-dataset} summarizes its composition and breaks it down by research problem.
The corpus spans 14 generator models including GPT-5.4, Claude Opus 4.6, Gemini-2.5-Flash, and several frontier preview models. Mechanisms include methods such as AdaEvolve, EvoX and GEPA. For our analysis, we focus on two problems that are representative of those commonly used to benchmark ADRS and for which we have the most comprehensive data coverage, 2D Knapsack and Polynomino Packing. Appendix~\ref{sec:appendix-problems} provides descriptions of the problems analyzed.

\subsection{Basin Structure (RQ1)}\label{sec:rq1}

\begin{figure}[!t]
\centering
\includegraphics[width=\linewidth]{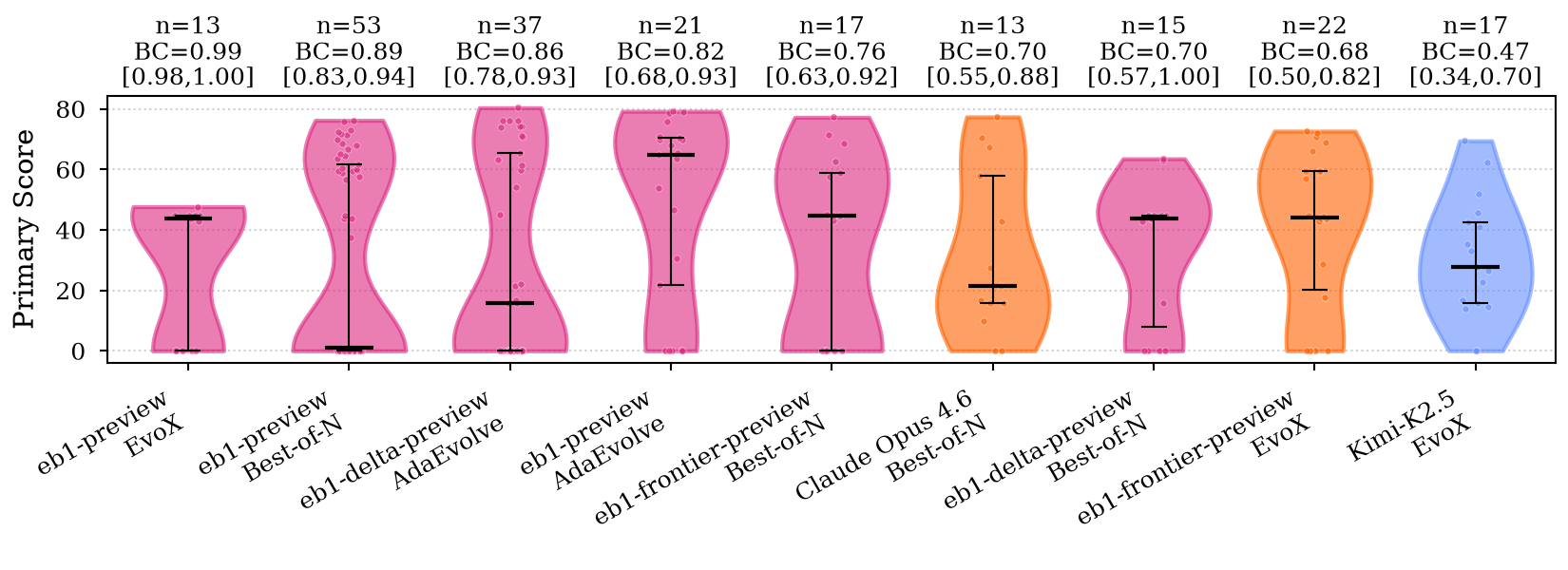}
\caption{Basin structure for Polynomino Packing for the 9 configurations with the most runs; the full set of 14 configurations with $n \geq 10$ appears in Figure~\ref{fig:rq1-polynomino-full} (Appendix~\ref{sec:appendix-rq1}). Color indicates classification: magenta = confirmed bimodal, orange = inconclusive due to  CI, blue = unimodal. Labels report sample size, point-estimate BC, and bootstrapped 95\% CI.}
\label{fig:rq1-polynomino}
\end{figure}

To detect whether optimization landscapes contain multiple basins, we analyze the distribution of final scores across runs with identical configurations.
Let $\mathcal{C}_\phi$ denote the set of campaigns sharing configuration $\phi$. For each campaign $c \in \mathcal{C}_\phi$, we compute the final score $s_c = \max_{x \in \text{candidates}(c)} \mathcal{A}(x)$.

To quantify whether the distribution of final scores exhibits multiple modes, we compute the bimodality coefficient:
\begin{equation}
\text{BC}_\phi = \frac{\gamma_\phi^2 + 1}{\kappa_\phi + 3}.
\label{eq:bimodality}
\end{equation}
where $\gamma_\phi$ is the skewness and $\kappa_\phi$ is the excess kurtosis of $\{s_c : c \in \mathcal{C}_\phi\}$. A value of $\text{BC}_\phi > 5/9 \approx 0.555$ suggests bimodality, while $\text{BC}_\phi \leq 5/9$ suggests unimodality.

\textbf{Empirical Analysis.}
We estimate BC and its 95\% bootstrap confidence interval (1,000 resamples) for each configuration with $n \geq 10$ runs, classifying a distribution as bimodal only when the CI lower bound exceeds $5/9$. Figure~\ref{fig:rq1-polynomino} shows the results for Polynomino Packing; the full Polynomino set and Bounded 2D Knapsack appear in Appendix~\ref{sec:appendix-rq1}.

\textbf{Polynomino Packing.} Of 14 configurations, 7 show confirmed bimodality spanning 4 models and all 3 mechanisms. Scores form a continuous distribution from 0 to approximately 82, suggesting multiple local optima rather than a binary outcome. For example, eb1-preview / Best-of-N ($n=53$, BC $= 0.89$, CI $[0.83, 0.94]$) ranges from 0 to 76, and eb1-delta-preview / AdaEvolve ($n=37$, BC $= 0.86$, CI $[0.78, 0.93]$) ranges from 0 to 80. Only one configuration is classified as unimodal.

\textbf{Bounded 2D Knapsack.} Of 10 configurations, 5 show confirmed bimodality spanning 3 models and all 3 mechanisms. Unlike Polynomino, the structure is binary: runs either fail completely (score $\approx 0$) or achieve near-ceiling results (score $\geq 99$), with few intermediate outcomes. For example, GPT-OSS-20B / Best-of-N ($n=13$, BC $= 0.95$, CI $[0.84, 1.00]$) and eb1-delta-preview / AdaEvolve ($n=18$, BC $= 0.96$, CI $[0.84, 1.00]$). Two configurations deterministically reach the top basin across all runs.

Across both problems, 12 of 24 qualifying configurations exhibit confirmed multimodality, spanning multiple models and all mechanisms, and the two problems reveal distinct basin geometries. This suggests the basin structure is inherent to the optimization landscape rather than an artifact of specific configurations. For practitioners, a single run does not characterize a configuration's capability; replication is necessary to determine whether a setup reliably reaches the best basin. Identifying which configurations exhibit high $\text{BC}_\phi$ guides resource allocation: unimodal configurations need fewer runs while multimodal ones benefit from replication.

\subsection{Stagnation Heuristics (RQ2)}\label{sec:rq2}
Detecting when ADRS runs have become trapped in a local optimum is a question for theory, system design, and practice alike: it shapes convergence analysis, is embedded in mechanisms (AdaEvolve's adaptive intensity, EvoX's no-improvement switch), and tells practitioners when to continue or reallocate compute. All invoke one intuition, that a run without improvement for $k$ iterations is unlikely to improve soon; we ask whether it is calibrated for ADRS.

We take the \emph{rule of three} as a tractable representative. Modeling each
iteration as an independent Bernoulli trial with improvement probability $p$,
$k$ consecutive non-improving iterations give the 95\% upper bound
$p \lesssim 3/k$; the probability of at least one improvement in the next $k/3$
iterations, $1-(1-p)^{k/3}$, evaluates at $p=3/k$ to $1-e^{-1}\approx 0.63$. Because $3/k$ is an upper bound on $p$, this makes $0.63$ an upper bound on the
escape rate. This rests on the independence
assumption, which any ADRS that conditions generation on accumulated run history
violates~\cite{ellis_dont_2026}.
We measure the consequence: where observed escape
rates fall relative to $0.63$, near it (escape remains likely, and stopping
risks forgoing improvement) or far below it (runs are more stuck than the
heuristic allows, and patience risks wasted compute), and how this varies
across problems and generators. Pooling runs is what makes
the comparison measurable.

Formally, let $s_t = \max_{i \leq t} \mathcal{A}(x_i)$ denote the running best score at iteration $t$. We define a \textit{stagnation episode} beginning at iteration $t_0$ where $s_t = s_{t_0}$ for $t \in [t_0, t_0 + k)$. At threshold $k$, the escape indicator $E_k = \mathbf{1}[\exists\, t \in W_k : s_t > s_{t_0}]$ records whether the running best improves within the prediction window $W_k = [t_0 + k,\; t_0 + k + \lfloor k/3 \rfloor)$, and the observed escape rate $\hat{p}_k = \frac{1}{N_k}\sum_{i=1}^{N_k} E_k^{(i)}$ averages it over the $N_k$ uncensored stagnation episodes (runs surviving past the prediction window).
We evaluate at thresholds $k \in \{5, 10, 15, \ldots, 50\}$ and compare $\hat{p}_k$ against the reference rate of 0.63.

For each campaign, we compute a rolling best score and stagnation count at each iteration. When stagnation crosses a threshold, we record whether improvement occurred in the subsequent $\lfloor k/3 \rfloor$ iterations, i.e., within $W_k$.

\begin{figure}[t]
\centering
\includegraphics[width=0.7\linewidth]{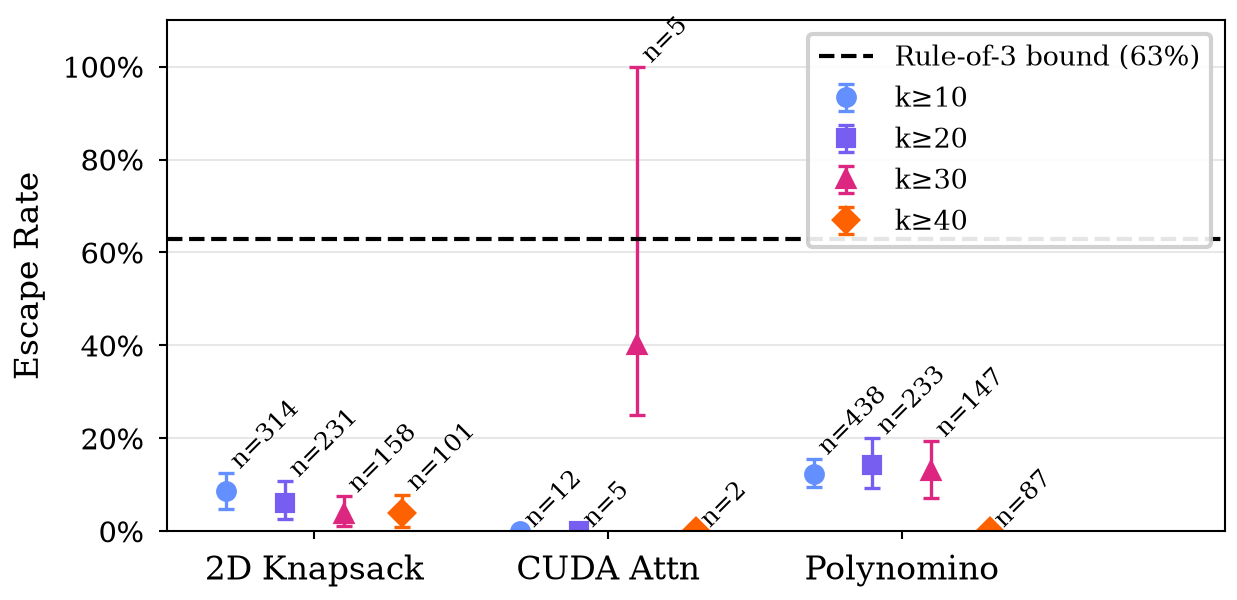}
\caption{Calibration of rule-of-3 across three problems. The dashed line at 63\% is the ceiling of the heuristic. Observed escape rates fall far below it.}
\label{fig:rq2-calibration}
\end{figure}

\textbf{Empirical Analysis.}
We measure rule-of-3 calibration across three problems: Bounded 2D Knapsack, CUDA Attention Kernel Optimization, and Polynomino Packing.

Aggregate results show miscalibration, as shown in Figure~\ref{fig:rq2-calibration}. Across the problems and stagnation thresholds, observed escape rates remain far below 63\%, typically under 15\%, falling to around 4\% at longer stagnation lengths. Rates below 63\% do not contradict the heuristic; what the pooled corpus adds is the magnitude of the gap, large enough that patience calibrated to it wastes most of the iterations it buys. To isolate the within-run effect of stagnation from between-run heterogeneity, we fit a mixed-effects logistic regression with a random intercept per run. The random intercept gives each run its own baseline escape probability, ensuring that runs which are inherently difficult do not confound the estimated trend. The resulting odds ratio is 0.91 per 10-step increase (95\% CI $[0.86, 0.96]$), meaning each additional 10 iterations of stagnation reduces the odds of escape by 9\%. Because the confidence interval lies entirely below 1 (the null value indicating no effect), the decline is statistically significant and not an artifact of run-level differences.

Problem-level heterogeneity is substantial. Bounded 2D Knapsack exhibits consistently low escape rates (4--9\%) aligning with its binary basin structure from RQ1, and Polynomino Packing shows moderate escape rates (13--14\%). CUDA Attention Kernel Optimization shows a striking anomaly at $k \geq 30$, where escape rate spikes to 40\%, but the large variance makes this observation inconclusive.

Mechanism differences are not clearly distinguishable (see Appendix~\ref{sec:appendix-rq2}). While point estimates suggest AdaEvolve sustains slightly higher escape rates for Polynomino Packing, the 95\% bootstrap confidence intervals overlap across all three mechanisms for both problems, suggesting that stagnation behavior is more heavily influenced by problem landscape structure than by mechanism choice.

For practitioners, escape is far rarer than the optimistic \emph{rule of three} allows; early termination may save compute with minimal loss of final score. For mechanism design, the persistent stagnation suggests that mechanisms may benefit from periodic context resets or diversity injection.

\subsection{Diversity and Context Composition (RQ3)}\label{sec:rq3}

We ask whether what a run explores, and what it shows the generator, relate to its outcomes. We measure code diversity by embedding candidate code with \textit{jina-embeddings-v2-base-code}~\citep{gunther_jina_2024} ($d = 768$; Appendix~\ref{app:performance}) and taking the average pairwise cosine distance $D(X)$ among the embeddings of a candidate set $X$ (Appendix~\ref{sec:appendix-rq3}).

\textbf{Early diversity.} At the run level, diversity among candidates from the first 25\% of iterations is at best a weak predictor of final score (Figure~\ref{fig:rq3-early}). It is weakly positive on Polynomino Packing ($\rho = 0.15$, $p = 0.011$), absent on Bounded 2D Knapsack ($\rho = 0.00$, $p = 0.943$), whose runs largely either reach the ceiling or fail outright, and inconclusive on the underpowered CUDA Attention ($\rho = -0.50$, $p = 0.667$); some configurations even reverse the Polynomino direction (EvoX, $\rho = -0.24$, $p = 0.030$).

\begin{figure}[!t]
\centering
\includegraphics[width=\linewidth]{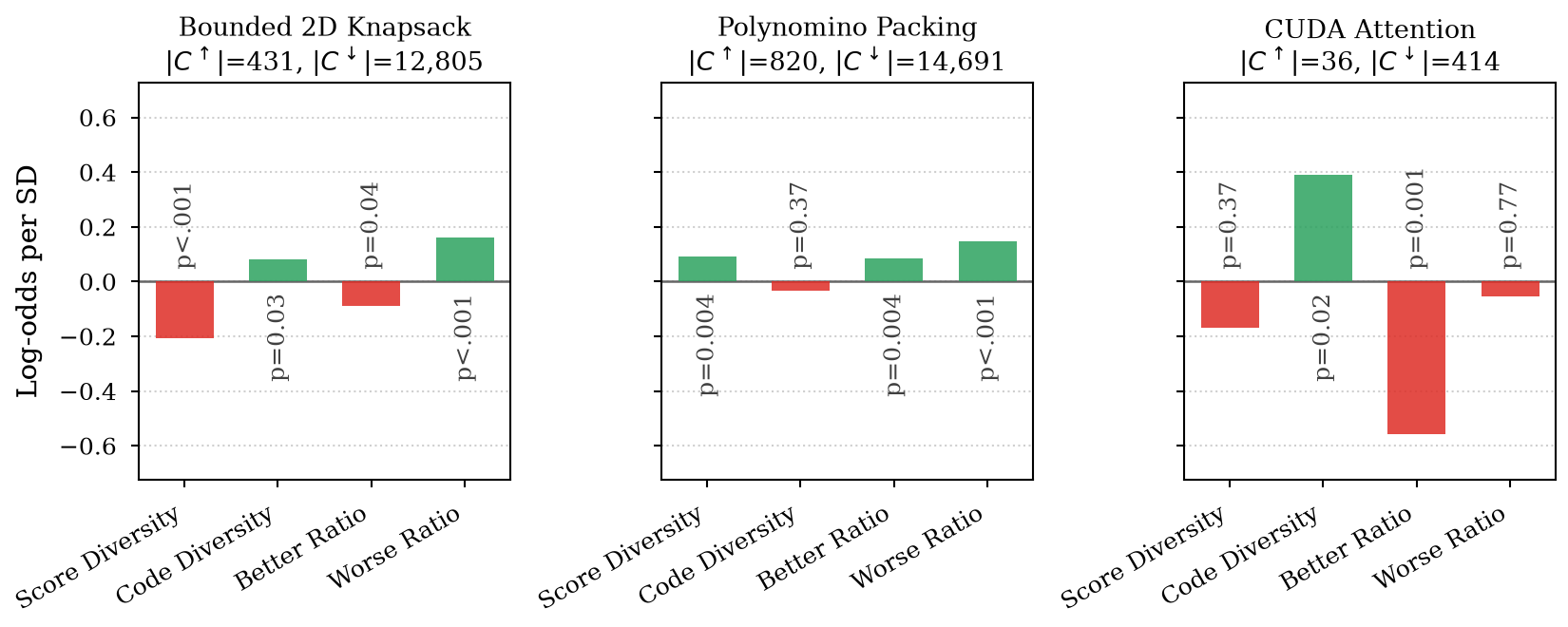}
\caption{Factor importance for predicting score improvement, by problem. Bars are log-odds of improvement per 1 SD of each factor; positive (green) indicates the factor is associated with more improvements, negative (red) with fewer. $|C^{\uparrow}|$, $|C^{\downarrow}|$ give the improving and non-improving candidate counts.}
\label{fig:rq4-factors}
\end{figure}

\textbf{Context composition.} We quantify how strongly each context-composition factor distinguishes improving from non-improving candidates. Let $C^{\uparrow} = \{x : \mathcal{A}(x) > s_{t-1}\}$ denote candidates that improved the running best score, and $C^{\downarrow} = \{x : \mathcal{A}(x) \leq s_{t-1}\}$ denote those that did not. For each candidate $x$ with parent $p$ and context set $\text{ctx}(x)$, we compute four factors: the fractions of context examples scoring above and below the parent (\emph{better} and \emph{worse context ratio}), the variance of context scores (\emph{score diversity}), and the diversity $D(\text{ctx}(x))$ of its context (\emph{code diversity}). Formal definitions are given in Appendix~\ref{sec:appendix-rq4-factors}.

Because candidates within a campaign share configuration and are serially dependent, we fit a mixed-effects logistic regression with a random intercept per campaign, as in RQ2, controlling for iteration phase. We standardize each factor to zero mean and unit variance before fitting, such that every coefficient is on a common scale, the change in log-odds of improvement per one standard deviation. Positive values indicate the factor is associated with more improvements, negative with fewer. Figure~\ref{fig:rq4-factors} shows these effects for each problem.

The three problems show distinct patterns. For Bounded 2D Knapsack, score diversity in context is the strongest signal and is negative ($\approx -0.2$), and better context ratio is also negative ($\approx -0.1$), whereas worse context ratio is positive ($\approx +0.15$). On this binary-basin problem, context that is more score-diverse coincides with fewer improvements, while a larger share of worse-scoring examples coincides with more.

Polynomino Packing shows a contrasting pattern in which the significant factors are all positive: worse context ratio ($\approx +0.15$), better context ratio ($\approx +0.09$), and score diversity ($\approx +0.09$) each raise the odds of improvement---showing the generator candidates across the quality spectrum, including failures, predicts better outcomes.

CUDA Attention reveals a third pattern: a higher fraction of better-scoring context examples is associated with fewer improvements (better context ratio $\approx -0.55$), while code diversity is positively associated ($\approx +0.4$). Score diversity and worse context ratio show no significant relationship therefore, it is difficult to decisively conclude a new pattern or a similar pattern to Bounded 2D Knapsack.

Disaggregating by model and mechanism jointly (Appendix~\ref{sec:appendix-rq4}) reveals heterogeneity beyond the aggregate patterns: although the overall trend is observed between the two problems, some model--mechanism combinations show significant positive effects and others significant negative effects.

Where significant, both early diversity and context composition are modest predictors, with log-odds shifts within roughly $\pm 0.2$ per SD in most cases, notable given that both are central design parameters in current ADRS mechanisms. Their direction is also problem dependent: early diversity carries signal only on Polynomino Packing, and context that shows cases to avoid helps on Bounded 2D Knapsack, whereas Polynomino Packing rewards examples across the quality spectrum. Fixed exploration biases and context selection strategies are therefore suboptimal, and practitioners should characterize the problem's landscape before committing to either.

\subsection{Budget Allocation (RQ4)}\label{sec:rq5}

We test a simple decision that pooled campaigns can inform: how to divide a fixed iteration budget $B$ as described in (\S\ref{sec:budget-allocation}) for the best performance using the same configuration. Any improvement is therefore attributable to allocation informed by pooled history rather than to more compute or a better configuration.

Comparisons are paired at the campaign level. For each stored campaign $c$, the baseline is the best score $c$ reached by iteration $B$. The policy-informed campaign starts from the same campaign $c$, but stops it early or caps it at $\lfloor B/n \rfloor$ iterations as $\pi$ dictates, and spends the remaining budget launching new campaigns under $\pi$. We simulate these later campaigns by drawing campaigns of the same configuration at random from ROAR, with replacement, and replaying their recorded trajectories; each draw stands in for an outcome the configuration could plausibly produce. The policy-informed score on $c$ is the mean over $T = 500$ such simulated executions, and its gain is the difference from the baseline. Within each tested configuration, every campaign is held out as the starting $c$. A configuration's gain is the mean over its held-out campaigns, and a problem's gain is the mean over all campaigns of its tested configurations. We report 95\% confidence intervals on this gain from a bootstrap over campaigns: within each configuration we resample the campaigns 1{,}000 times with replacement (small-pool caveat in Appendix~\ref{sec:appendix-rq5}), refit $\pi$ on each resample, and score the campaigns left out of it, so the intervals reflect the uncertainty of the whole fit-and-score procedure.

\textbf{Empirical Analysis.}
Among the two problems with at least 50 recorded iterations across, 17 configurations hold at least 8 campaigns each. A policy can only improve or regress on a configuration whose best scores vary across campaigns; twelve do and are tested (cutoff in Appendix~\ref{sec:appendix-rq5}). Table~\ref{tab:rq5-per-problem} gives the aggregate result; the per-configuration breakdown is in Appendix~\ref{sec:appendix-rq5}.

\begin{table}[t]
\centering
\footnotesize
\caption{RQ4: gain of the ROAR budget policy over a single 50-iteration campaign, per problem. Campaigns is the number of campaigns in the included configurations; each is held out once and scored against the policy fit on the others. Gain is the mean difference in best score. Intervals are 95\% bootstrap confidence intervals over campaigns, with the policy refit on each resample.}
\label{tab:rq5-per-problem}
\setlength{\tabcolsep}{4pt}
\begin{tabular}{lrrrr}
\toprule
\textbf{Problem} & \textbf{Cfgs} & \textbf{Campaigns} & \textbf{Gain} & \textbf{95\% CI} \\
\midrule
Polynomino  & 9 & 115 & $+18.9$ & $[+11.6, +25.6]$ \\
Knapsack    & 3 & 37  & $+16.9$ & $[-1.0, +33.6]$ \\
\bottomrule
\end{tabular}
\end{table}

On Polynomino Packing the policy is worth $+18.9$ points of final score $[+11.6, +25.6]$ across 115 campaigns in 9 configurations, at an identical budget. Bounded 2D Knapsack is positive but not significant, $+16.9$ $[-1.0, +33.6]$ over 37 campaigns in 3 configurations; with only three configurations that retain headroom, this problem is underpowered rather than shown to be unaffected.

Nonetheless, the gain trends positive in 11 of the 12 tested configurations. The fitted policy itself is not shared, for example, the fitted $k_{\text{stop}}$ ranges from 2 to 22, even between configurations with the same problem. The threshold therefore has to be estimated per configuration, which is what pooled campaigns make possible.

\section{Conclusion}

We presented ROAR, the first general data aggregation infrastructure for ADRS, built on a unified relational schema with a parser layer that translates each framework's native outputs into a common representation. Our findings cohere into a general approach for ADRS design that departs from current practice. First, because runs may settle into local optima earlier and more persistently than common stopping heuristics account for (\S\ref{sec:rq2}), extended single runs may yield diminishing returns; the same compute can be better allocated across several shorter, replicated runs, which have the added benefit of sampling distinct regions of the landscape (\S\ref{sec:rq1}). We test this result in addition to asserting it: reallocating a fixed budget according to a early stopping and replication policy fitted from pooled history is worth $+18.9$ points of final score on Polynomino Packing and yields a positive gain in 11 of 12 configurations, at no additional compute (\S\ref{sec:rq5}).  Third, because the value of solution diversity and context composition varies across problems (\S\ref{sec:rq3}), exploration and exploitation should be tuned to the characteristics of the problem landscape rather than fixed a priori.
Each step presumes something a single team rarely has: enough runs, across enough problems or ADRS, to know where a problem sits. Pooling results is what turns these observations from post hoc findings into design-time guidance. Despite covering more than 900 runs across 7 problems and 14 models, the dataset remains limited relative to the space of ADRS configurations and tasks, and extending coverage will require community adoption; as it grows, it becomes feasible to build data-driven configuration systems that recommend mechanism choice and context-selection strategy for new problems.

%
\subsection*{AI use statement}

The ADRS campaigns analyzed in this work are produced by LLM-driven
systems; they are the object of study, not a tool used to conduct the
research. In conducting the research, we used generative AI tools to
create or edit software code, create or modify scientific figures, draft
parts of this paper, and edit the paper to improve readability. We did not
use generative AI tools to generate data, develop theoretical models or
conceptual frameworks, propose or refine hypotheses, design or provide
feedback on research methodology or experiments, clean or reformat data
sets, or interpret results. Formulating or proving mathematical claims and
translation are not applicable to this work. All AI-generated code was
verified and tested by the authors, and all AI-assisted text was checked
by the authors. We take responsibility for the final content of this work,
including text, claims, or artifacts produced with the aid of generative
AI.

\subsection*{Reproducibility statement}

Code and data are released at the link in the footnote of
Section~\ref{sec:intro}. The release contains the ROAR database schema
and ingestion code, the full dataset of campaigns, both parsed and raw.
Section~\ref{sec:method} describes how the components fit together: the
schema that normalizes heterogeneous ADRS outputs
(Appendix~\ref{sec:appendix_schema}), the ingestion API and parsers, the
analytical pipeline, and the budget allocation policy fit (Section~\ref{sec:budget-allocation},
Appendix~\ref{sec:appendix-rq5}). The dataset composition, generator
models, and providers are listed in Appendix~\ref{sec:appendix-dataset},
and the benchmark problems in Appendix~\ref{sec:appendix-problems}.
Section~\ref{sec:experiments} states the inclusion criteria and
statistical tests for each research question. The
hardware used is given in Appendix~\ref{app:performance}.

\bibliography{iclr2027}
\bibliographystyle{iclr2027_conference}

\appendix

\section{Extended Related Work}
\label{sec:appendix-related-work}

ROAR connects two lines of work: AI-driven research systems (ADRS) design, as a new emerging subfield, and ADRS behavior analysis.

\textbf{AI-Driven Research Systems}. Building on earlier works that cast a language model as a mutation operator \citep{lehman2022evolutionlargemodels,li2022alphacode}, early ADRS such as FunSearch~\citep{Romera-Paredes2024} and  AlphaEvolve~\citep{novikov_alphaevolve_2025} placed a language model inside an evolutionary loop, paired with a systematic evaluator. They demonstrated astonishing breakthroughs, such as the first improvement on Strassen's algorithm for four-by-four complex matrix multiplication in 56 years.
Open source frameworks such as OpenEvolve \citep{openevolve}, ShinkaEvolve \citep{lange_shinkaevolve_2025}, and GEPA \citep{agrawal_gepa_2026} followed; each using a fixed search strategy and developing how candidate solutions to a given problem are selected and varied.

Later work introduced frameworks that adapt search parameters or algorithms during runtime. For instance, AdaEvolve \citep{cemri_adaevolve_2026}, EvoX \citep{liu_evox_2026}, and SkyDiscover \citep{liu_skydiscover_2026} all reallocate compute toward whichever candidate pools are still improving rather than committing to one strategy in advance. These frameworks rest on the principle that a population of candidates is ranked by score and evolved over time.

An emerging family of frameworks moves away from maintaining explicit candidate pools and instead treats the system as an autonomous researcher. These approaches coordinate specialized agents for reasoning, experimentation, and analysis, often using persistent summaries or archives to maintain continuity across long-horizon investigations. Representative systems include Glia \citep{hamadanian2026glia}, Engram \citep{karimi2026engram}, and CORAL \citep{qu2026coral}.

These increasingly diverse frameworks have already been applied to practical systems problems, including request routing, scheduling, auto-scaling, and KV-cache reuse~\cite{cheng_barbarians_2025}.
Each ADRS records run history in significantly different ways, hampering cross-comparison for any target problem.
This paper's proposed data aggregation infrastructure addressing this gap, where a generic experiment tracker such as MLflow \citep{zaharia_mlflow_2018} is insufficient, since it leaves naming conventions and semantic abstractions up to each team.

\textbf{Analyzing ADRS Behavior}.
A growing body of work analyzes the behavior of ADRS. One line of research examines prompt construction, evaluator design, and experimental harness roles\citep{cheng_barbarians_2025,cheng_let_2025,tanveer2026levi,metaharness2026}, including a domain-specific study of ADRS for database systems~\citep{cheng2026adrsdatabases}. These studies seek to improve the applicability, reliability, and performance of ADRS. A complementary line of work develops abstractions for explaining differences in ADRS behavior, as exemplified by the GAMBLe framework~\citep{ellis_dont_2026}.

A common limitation of these studies is that their analyses rely primarily on runs generated by the authors under controlled experimental conditions. Broader access to runs from multiple systems and research groups could support more comprehensive analyses, which motivates the ROAR platform. However, aggregating externally produced runs introduces a new challenge: evaluation procedures and metrics may differ substantially across studies. ROAR addresses this gap by introducing analytical paradigms that can extract useful insights from these heterogeneous runs.

\FloatBarrier

\section{Dataset Details}
\label{sec:appendix-dataset}

Table~\ref{tab:dataset} summarizes the scale and composition of the currently collected data, and Table~\ref{tab:problem-coverage} provides a detailed breakdown by research problem.

\begin{table}[h]
\centering
\caption{Dataset summary statistics.}
\label{tab:dataset}
\begin{tabular}{lr}
\toprule
\textbf{Metric} & \textbf{Count} \\
\midrule
Campaigns (runs) & 928 \\
Candidates & 48,210 \\
\midrule
Distinct generators & 14 \\
Distinct mechanisms & 6 \\
\bottomrule
\end{tabular}
\end{table}

\begin{table}[ht]
\centering
\small
\caption{Problem coverage in the dataset. The three combinatorial problems dominate the corpus; CUDA kernel problems provide signal on continuous landscapes; single-campaign problems are excluded from statistical analyses.}
\label{tab:problem-coverage}
\begin{tabular}{lrrrr}
\toprule
\textbf{Problem} & \textbf{Runs} & \textbf{Cands} & \textbf{Mechs} & \textbf{Models} \\
\midrule
Palindrome Ham.\ Path & 290 & 17,181 & 3 & 10 \\
Polynomino Packing & 353 & 16,293 & 3 & 12 \\
Bounded 2D Knapsack & 271 & 13,857 & 3 & 12 \\
CUDA Attention & 3 & 481 & 2 & --- \\
CUDA Layer Norm & 4 & 269 & 2 & --- \\
Dispatcher Cold Path & 6 & 68 & 5 & 1 \\
KV Cache Eviction & 1 & 61 & 1 & 1 \\
\midrule
\textbf{Total} & \textbf{928} & \textbf{48,210} & \textbf{19} & \textbf{36} \\
\bottomrule
\end{tabular}
\end{table}

\medskip

Table~\ref{tab:model-names} lists all generator models and their providers.

\begin{table}[h]
\centering
\caption{Generator models and their providers.}
\label{tab:model-names}
\begin{tabular}{ll}
\toprule
\textbf{Model} & \textbf{Provider} \\
\midrule
GPT-5.4 & OpenAI \\
GPT-5-mini & OpenAI \\
GPT-OSS-20B & OpenAI \\
Kimi-K2.5 & Moonshot AI \\
Claude Opus~4.6 & Anthropic \\
Claude Sonnet~4.6 & Anthropic \\
eb1 & Stanford \\
eb1-preview & Stanford \\
eb1-delta-preview & Stanford \\
eb1-frontier-preview & Stanford \\
eb1-pro & Stanford \\
Gemini-2.5-Flash & Google \\
Gemini-3-Flash-Preview & Google \\
Qwen-3.5-9B & Alibaba \\
\bottomrule
\end{tabular}
\end{table}

\FloatBarrier

\clearpage 
\section{ADRS Database Schema}
\label{sec:appendix_schema}

The Agentic Discovery Research Storage (ADRS) database uses a PostgreSQL schema with six core tables. This appendix provides the complete schema reference.

\setlength{\intextsep}{8pt plus 2pt minus 2pt} 

\subsection{Systems}

Identifies the research framework that produced the data.

\begin{table}[h]
\centering
\footnotesize
\begin{tabular}{lll}
\toprule
\textbf{Column} & \textbf{Type} & \textbf{Constraint} \\
\midrule
id & UUID & PK \\
name & TEXT & NOT NULL \\
version & TEXT & \\
\bottomrule
\end{tabular}
\caption{Systems table schema}
\label{tab:schema_systems}
\end{table}

\FloatBarrier

\subsection{Campaigns}

A campaign represents one top-level research run, linked to a system.

\begin{table}[h]
\centering
\footnotesize
\begin{tabular}{@{}p{0.45\linewidth}p{0.27\linewidth}p{0.2\linewidth}@{}}
\toprule
\textbf{Column} & \textbf{Type} & \textbf{Constraint} \\
\midrule
id & UUID & PK \\
system\_id & UUID & FK $\rightarrow$ systems \\
author & TEXT & NOT NULL \\
name & TEXT & \\
research\_question & TEXT & \\
research\_question\_embedding & vector(384) & \\
started\_at & TIMESTAMP & \\
ended\_at & TIMESTAMP & \\
config\_used & JSONB & \\
algorithm\_used & TEXT & \\
models\_used & TEXT[] & \\
total\_cost\_usd & DOUBLE & \\
total\_tokens & BIGINT & \\
final\_summary & TEXT & \\
final\_metrics & JSONB & \\
evaluator\_setup & JSONB & \\
\bottomrule
\end{tabular}
\caption{Campaigns table schema.}
\label{tab:schema_campaigns}
\end{table}

\FloatBarrier

\subsection{Candidates}

A candidate is the central object---the thing the system proposed. Each candidate belongs to a campaign and has an \texttt{iteration\_index} indicating which step of exploration it came from.

\begin{table}[h]
\centering
\footnotesize
\begin{tabular}{@{}p{0.5\linewidth}p{0.22\linewidth}p{0.22\linewidth}@{}}
\toprule
\textbf{Column} & \textbf{Type} & \textbf{Constraint} \\
\midrule
id & UUID & PK \\
campaign\_id & UUID & FK $\rightarrow$ campaigns \\
iteration\_index & INTEGER & NOT NULL \\
external\_id & TEXT & UNIQUE \\
candidate\_type & TEXT & \\
created\_at & TIMESTAMP & \\
solution\_summary & TEXT & \\
solution\_summary\_embedding & vector(384) & \\
direct\_code\_embedding & vector(768) & \\
context\_code\_diversity & DOUBLE & \\
\bottomrule
\end{tabular}
\caption{Candidates table schema.}
\label{tab:schema_candidates}
\end{table}

\FloatBarrier

\subsection{Candidate Edges}

Edges describe relationships between candidates, enabling lineage tracking and graph visualizations.

\begin{table}[h]
\centering
\footnotesize
\begin{tabular}{lll}
\toprule
\textbf{Column} & \textbf{Type} & \textbf{Constraint} \\
\midrule
id & UUID & PK \\
source\_candidate\_id & UUID & FK $\rightarrow$ candidates \\
target\_candidate\_id & UUID & FK $\rightarrow$ candidates \\
edge\_type & TEXT & DEFAULT 'parent' \\
\bottomrule
\end{tabular}
\caption{Candidate edges table schema. Edge types: \texttt{parent} (mutation lineage) and \texttt{context}.}
\label{tab:schema_edges}
\end{table}

\FloatBarrier

\subsection{Measurements}

A measurement is a named value associated with a candidate. All evaluation outcomes, metrics, and observations are stored as simple name-value pairs.

\begin{table}[h]
\centering
\footnotesize
\begin{tabular}{lll}
\toprule
\textbf{Column} & \textbf{Type} & \textbf{Constraint} \\
\midrule
id & UUID & PK \\
candidate\_id & UUID & FK $\rightarrow$ candidates \\
name & TEXT & NOT NULL \\
value & TEXT & NOT NULL \\
\bottomrule
\end{tabular}
\caption{Measurements table schema}
\label{tab:schema_measurements}
\end{table}

\FloatBarrier

\subsection{Artifacts}

An artifact is a reference to a file stored on disk. The database stores only metadata for tracking and integrity verification.

\begin{table}[h]
\centering
\footnotesize
\begin{tabular}{lll}
\toprule
\textbf{Column} & \textbf{Type} & \textbf{Constraint} \\
\midrule
id & UUID & PK \\
campaign\_id & UUID & FK $\rightarrow$ campaigns \\
candidate\_id & UUID & FK $\rightarrow$ candidates (nullable) \\
iteration\_index & INTEGER & \\
uri & TEXT & NOT NULL \\
content\_hash & TEXT & \\
size\_bytes & BIGINT & \\
mime\_type & TEXT & \\
\bottomrule
\end{tabular}
\caption{Artifacts table schema}
\label{tab:schema_artifacts}
\end{table}

\FloatBarrier

\subsection{Relationships}

The foreign key relationships are:

\begin{itemize}
    \item \textbf{systems} $\xrightarrow{1:N}$ \textbf{campaigns}: One system produces many campaigns
    \item \textbf{campaigns} $\xrightarrow{1:N}$ \textbf{candidates}: One campaign contains many candidates
    \item \textbf{campaigns} $\xrightarrow{1:N}$ \textbf{artifacts}: Campaign-level artifacts (config, logs)
    \item \textbf{candidates} $\xrightarrow{1:N}$ \textbf{measurements}: Each candidate has multiple measurements
    \item \textbf{candidates} $\xrightarrow{1:N}$ \textbf{candidate\_edges}: Lineage graph edges
    \item \textbf{candidates} $\xrightarrow{0..1:N}$ \textbf{artifacts}: Candidate-specific artifacts (optional)
\end{itemize}

\setlength{\intextsep}{12pt plus 2pt minus 2pt}

\section{System Performance}
\label{app:performance}

We report wall-clock performance for the three main data-processing stages: parsing, database insertion, and embedding generation for a subset of campaigns in the database.
This subset represents 914 campaigns (47,331 candidates, 322,595 measurements, 55,028 artifacts, 221,062 edges) from 59.4~GB of raw files.

\subsection{Hardware Configuration}

\paragraph{Parsing and Insertion.} Conducted in a development container with 8 parallel workers (Fedora Linux, Python 3.12, PostgreSQL with pgvector 0.8.2) allocated 8 vCPUs and 8~GB memory.

\paragraph{Embedding.} Generated on a compute cluster node with an NVIDIA H100 80GB HBM3 GPU, CUDA 12.4, and 64~GB system memory.

\subsection{Embedding Procedure}

We partition source code into overlapping chunks of 2048 tokens with 256-token overlap, embed each chunk independently using \textit{jina-embeddings-v2-base-code}, and average the resulting vectors to obtain a single embedding per candidate. This chunking process addresses the issue of large programs that exceed the context window of the model. Embeddings are computed via GPU-accelerated batch processing and stored directly in the database via pgvector for performant analysis queries.

\subsection{Performance Results}

Table~\ref{tab:performance} summarizes throughput for each processing stage.

\begin{table}[ht]
\centering
\small
\caption{Data processing performance across pipeline stages.}
\label{tab:performance}
\begin{tabular}{lrrr}
\toprule
\textbf{Stage} & \textbf{Records} & \textbf{Time (s)} & \textbf{Throughput} \\
\midrule
Parse & 914 & 278.6 & 3.3 camp/s \\
Insert & 914 & 16.6 & 55.1 camp/s \\
Embed & 48,210 & 2,418 & 19.9 rec/s \\
\bottomrule
\end{tabular}
\end{table}

\section{Problem Descriptions}
\label{sec:appendix-problems}

\paragraph{Bounded 2D Knapsack.}
Given a rectangular container and a set of items with integer dimensions, weights, and values,
pack a subset of items without rotation or overlap to maximize total value subject to a weight
capacity constraint.

\paragraph{Polynomino Packing.}
Given a fixed grid and a set of polyomino pieces, place as many pieces as possible without
overlap to maximize grid coverage.

\paragraph{CUDA Attention Kernel Optimization.}
Given a fixed kernel signature for bf16 scaled dot-product attention, optimize the raw CUDA
implementation to maximize runtime performance over a reference implementation on NVIDIA H100.

\FloatBarrier
\section{Additional RQ1 Figures}
\label{sec:appendix-rq1}
\suppressfloats[t]

This section presents the full configuration-level basin-structure figures for RQ1 (\S\ref{sec:rq1}): the complete Polynomino Packing set, of which Figure~\ref{fig:rq1-polynomino} in the main text shows a subset, and Bounded 2D Knapsack.

\begin{figure}[!htbp]
\centering
\includegraphics[width=\linewidth]{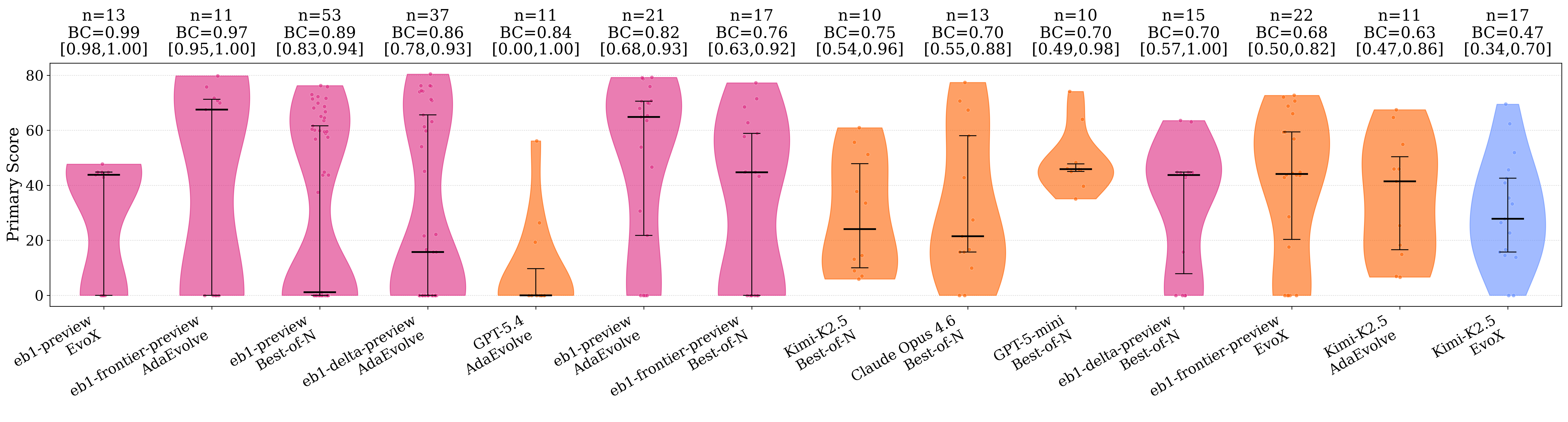}
\caption{Basin structure for Polynomino Packing for all 14 configurations with $n \geq 10$ runs; 16 configurations with fewer runs omitted. Color indicates classification: magenta = confirmed bimodal, orange = CI inconclusive, blue = unimodal. Labels report sample size, point-estimate BC, and bootstrapped 95\% CI. Of 14 configurations, 7 are confirmed bimodal, 6 are inconclusive, and 1 is unimodal.}
\label{fig:rq1-polynomino-full}
\end{figure}

\begin{figure}[!htbp]
\centering
\includegraphics[width=\linewidth]{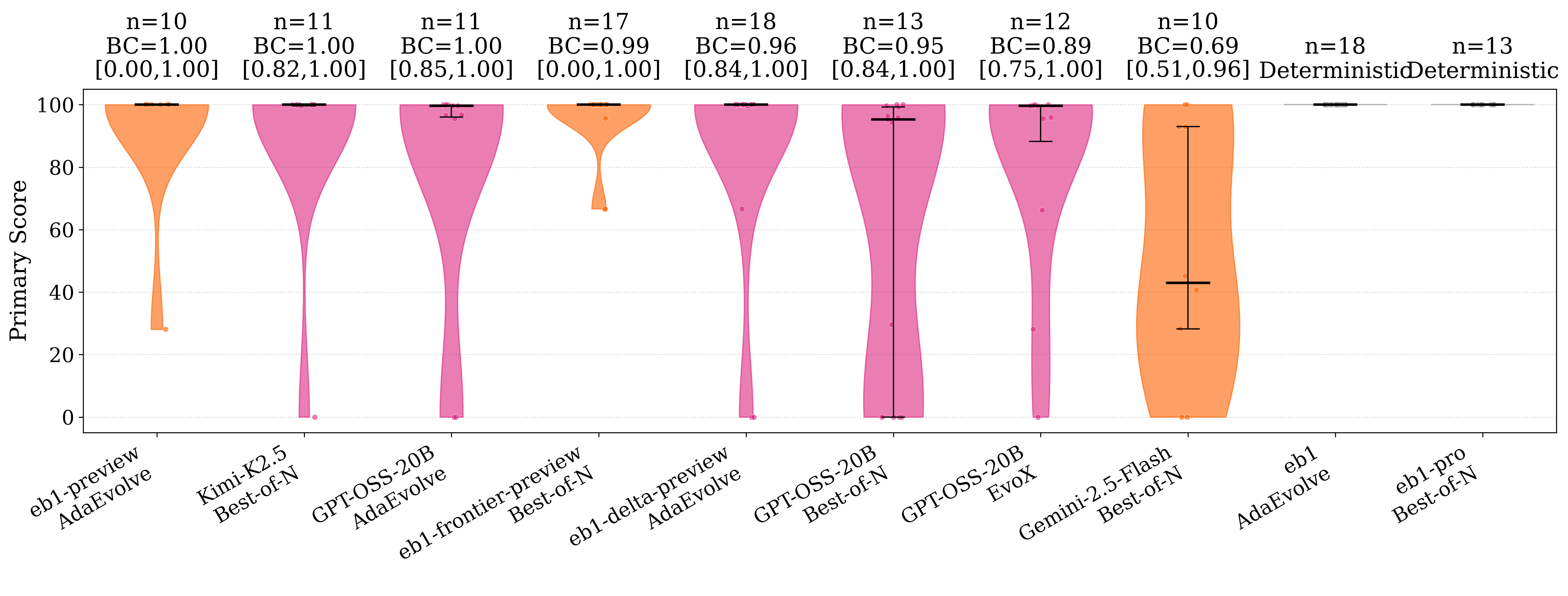}
\caption{Basin structure for Bounded 2D Knapsack at the configuration level. Each violin represents runs from a single configuration. Color indicates classification: magenta = confirmed bimodal, orange = CI inconclusive, gray = deterministic. Of 10 configurations, 5 are confirmed bimodal, 3 are inconclusive, and 2 are deterministic.}
\label{fig:rq1-knapsack-config}
\end{figure}

\FloatBarrier

\section{Additional RQ2 Figures}
\label{sec:appendix-rq2}

This section presents additional visualizations for RQ2 (Stagnation Heuristics) showing escape rate breakdowns by mechanism and model. Shaded bands indicate 95\% bootstrap confidence intervals computed by resampling at the campaign (run) level to respect within-run correlation.

\begin{figure}[h]
\centering
\includegraphics[width=0.75\linewidth]{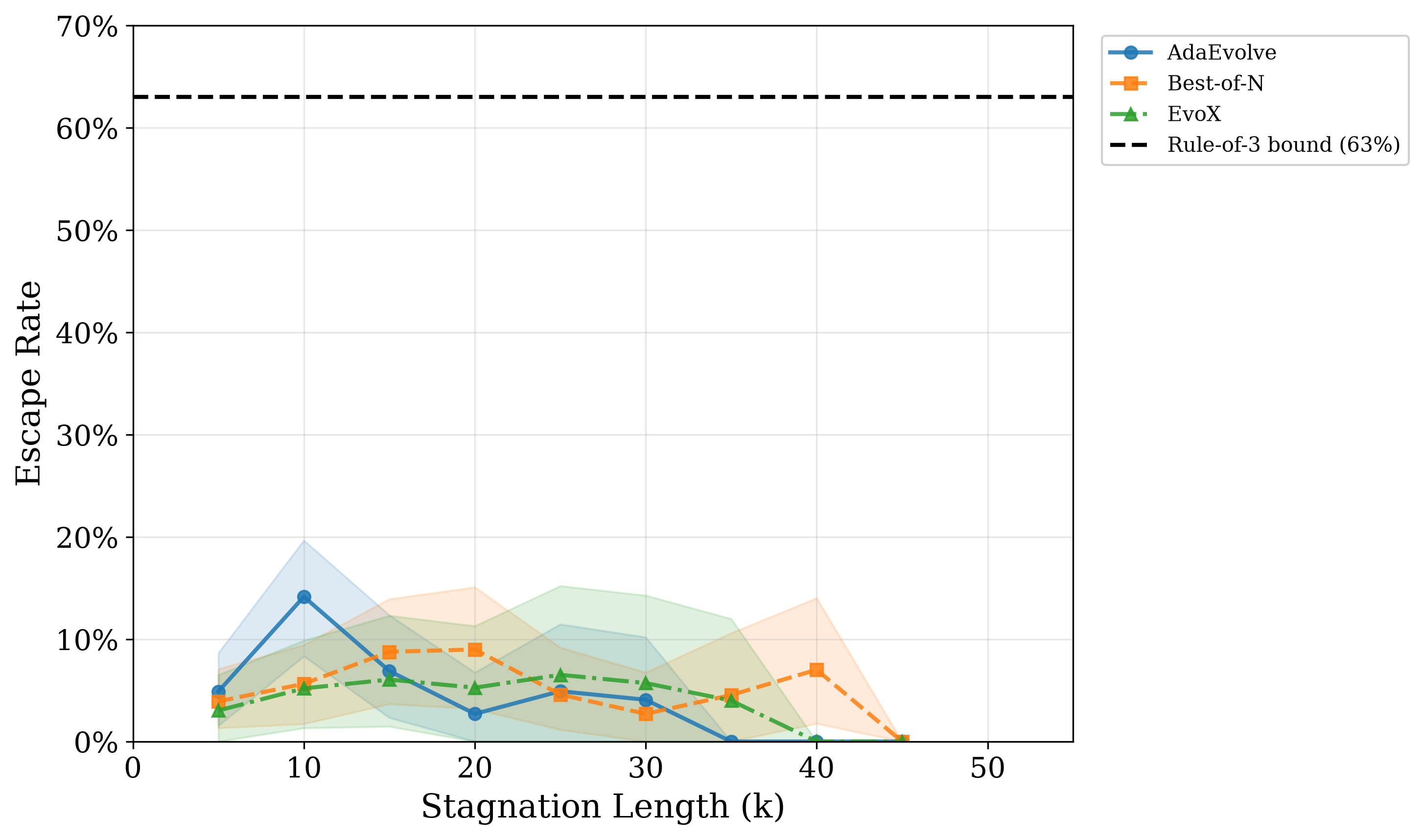}
\caption{Escape rate vs.\ stagnation length by mechanism for Bounded 2D Knapsack.}
\label{fig:rq2-knapsack-by-algorithm}
\end{figure}

\begin{figure}[h]
\centering
\includegraphics[width=0.75\linewidth]{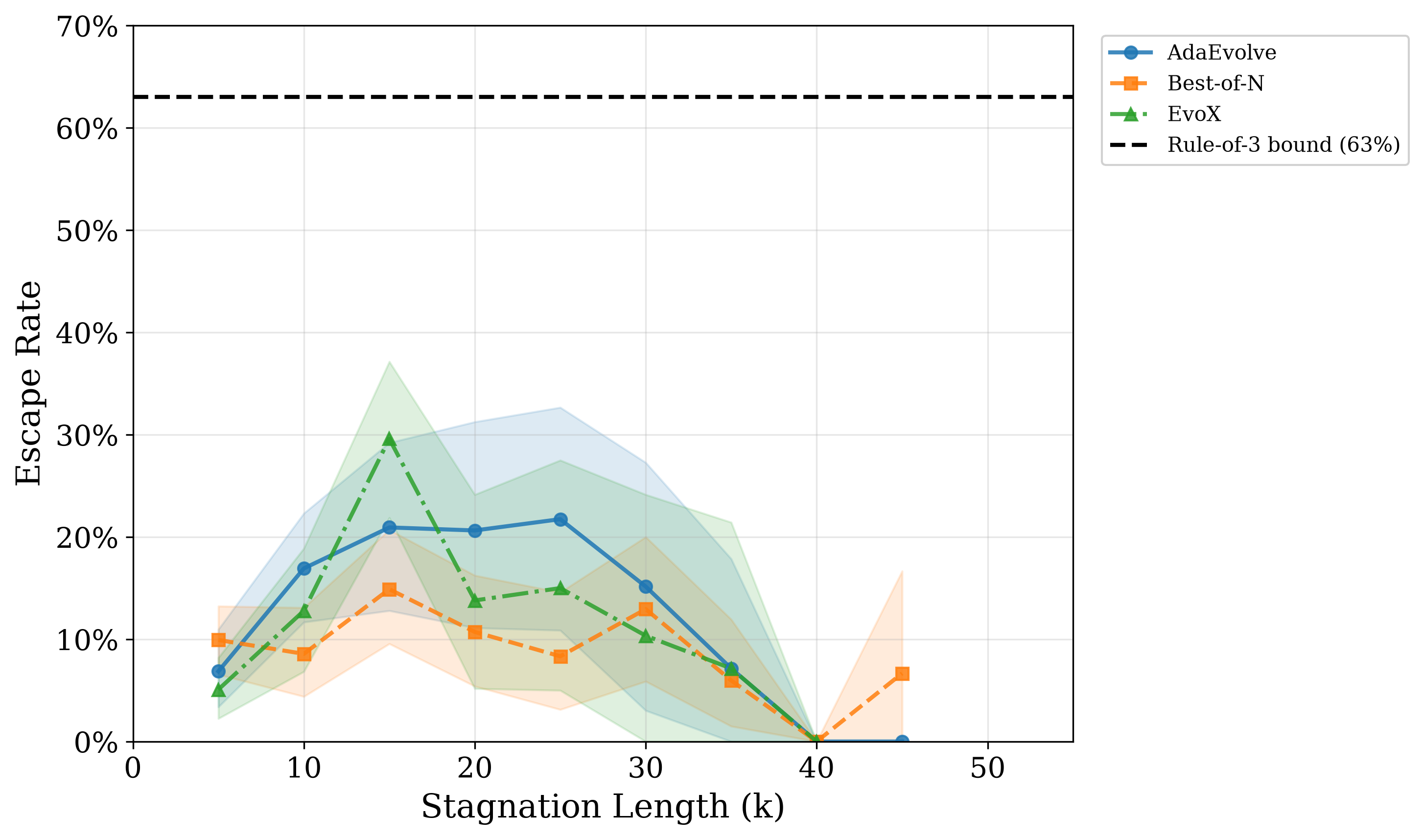}
\caption{Escape rate vs.\ stagnation length by mechanism for Polynomino Packing.}
\label{fig:rq2-by-algorithm}
\end{figure}

\FloatBarrier

\section{Additional RQ3 Details: Early Diversity}
\label{sec:appendix-rq3}

\subsection{Code Diversity}

To quantify diversity, we embed candidate code into a continuous vector space using \textit{jina-embeddings-v2-base-code}~\citep{gunther_jina_2024}, a code-specialized model producing $d = 768$ dimensional embeddings (details in Appendix~\ref{app:performance}). Each candidate $x$ with source code $\text{code}(x)$ is embedded as $e_x = f(\text{code}(x)) \in \mathbb{R}^{768}$.

Given a set of candidates $X = \{x_1, \ldots, x_n\}$ with embeddings $\{e_1, \ldots, e_n\}$, we define campaign-level diversity as the average pairwise cosine distance:
\begin{equation}
D(X) = \frac{2}{n(n-1)} \sum_{i < j} \left(1 - \frac{e_i \cdot e_j}{\|e_i\| \|e_j\|}\right).
\label{eq:diversity}
\end{equation}

This metric ranges from 0 (identical candidates) to a theoretical maximum of 2 (anti-correlated embeddings), though in practice code embeddings fall below 1. For systems that provide context examples to the generator, we also compute context diversity $D_{\text{ctx}}(x) = D(\text{ctx}(x))$ where $\text{ctx}(x)$ denotes the candidates shown when generating $x$.

\subsection{Early Diversity and Final Score}

\begin{figure}[h]
\centering
\includegraphics[width=0.32\linewidth]{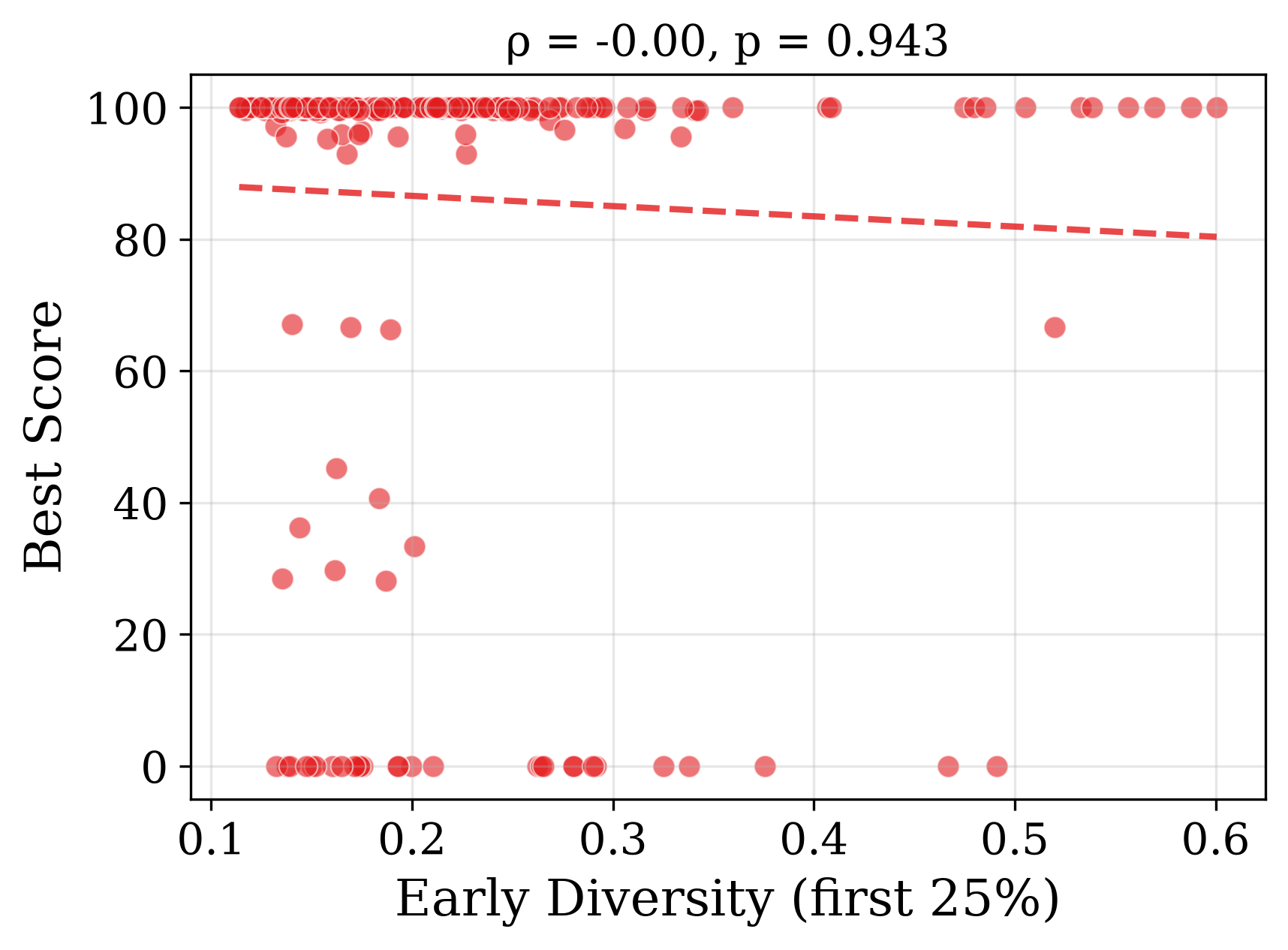}\hfill
\includegraphics[width=0.32\linewidth]{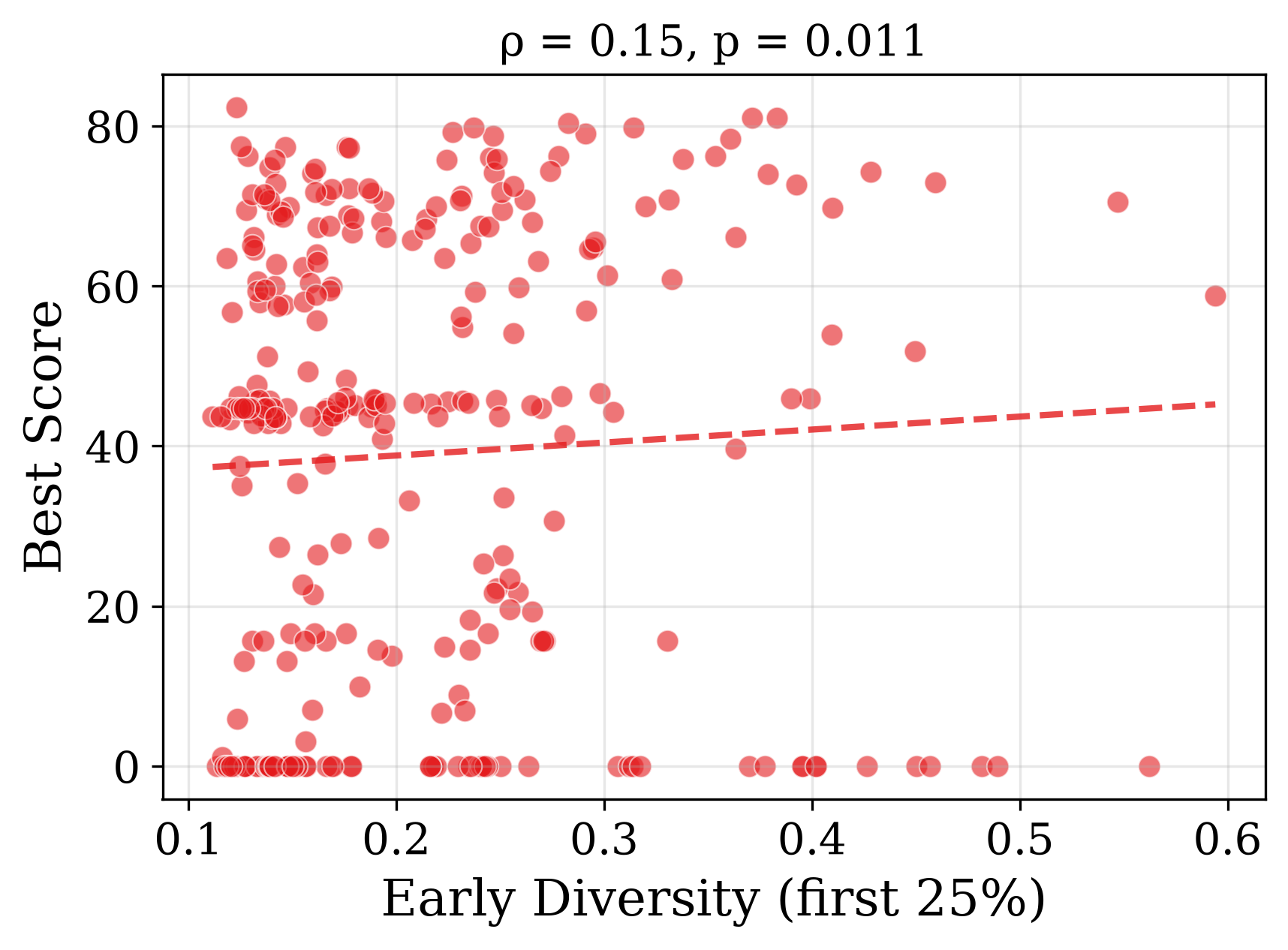}\hfill
\includegraphics[width=0.32\linewidth]{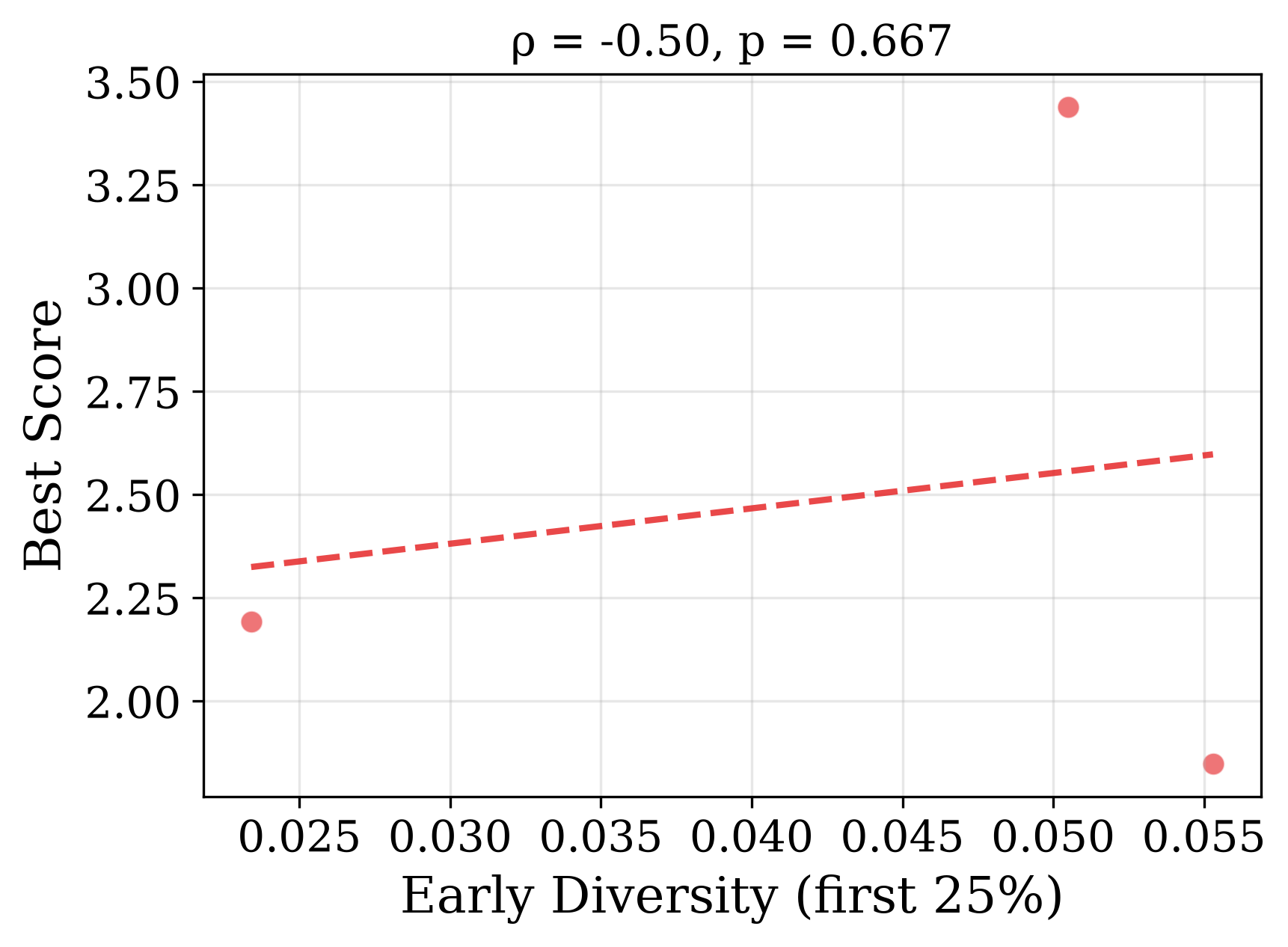}
\caption{Early diversity (first 25\% of iterations) vs.\ final score for Bounded 2D Knapsack (left), Polynomino Packing (center), and CUDA Attention (right). Only Polynomino Packing shows a significant relationship, which is positive; Bounded 2D Knapsack shows no early-diversity signal, and CUDA Attention is inconclusive given the limited sample size. Each panel title reports the Spearman rank correlation $\rho$ between early diversity and final score and its two-sided $p$-value.}
\label{fig:rq3-early}
\end{figure}

We analyze diversity among candidates from the first 25\% of iterations (Figure~\ref{fig:rq3-early}). Restricting to early iterations better isolates whether initial exploration predicts final performance. We quantify each diversity--score association with Spearman's rank correlation coefficient $\rho$ and its two-sided $p$-value, which captures monotonic relationships without assuming linearity. Polynomino Packing shows a weak but significant positive correlation ($\rho = 0.15$, $p = 0.011$), consistent with its continuous multi-optima landscape where diverse exploration helps to locate better local optima. Bounded 2D Knapsack shows no relationship ($\rho = -0.00$, $p = 0.943$): on its binary landscape, where runs largely either reach the ceiling or fail outright, early diversity carries little signal about the final outcome. CUDA Attention is too underpowered to draw a statistically significant conclusion ($\rho = -0.50$, $p = 0.667$).

\begin{figure}[h]
\centering
\includegraphics[width=0.7\linewidth]{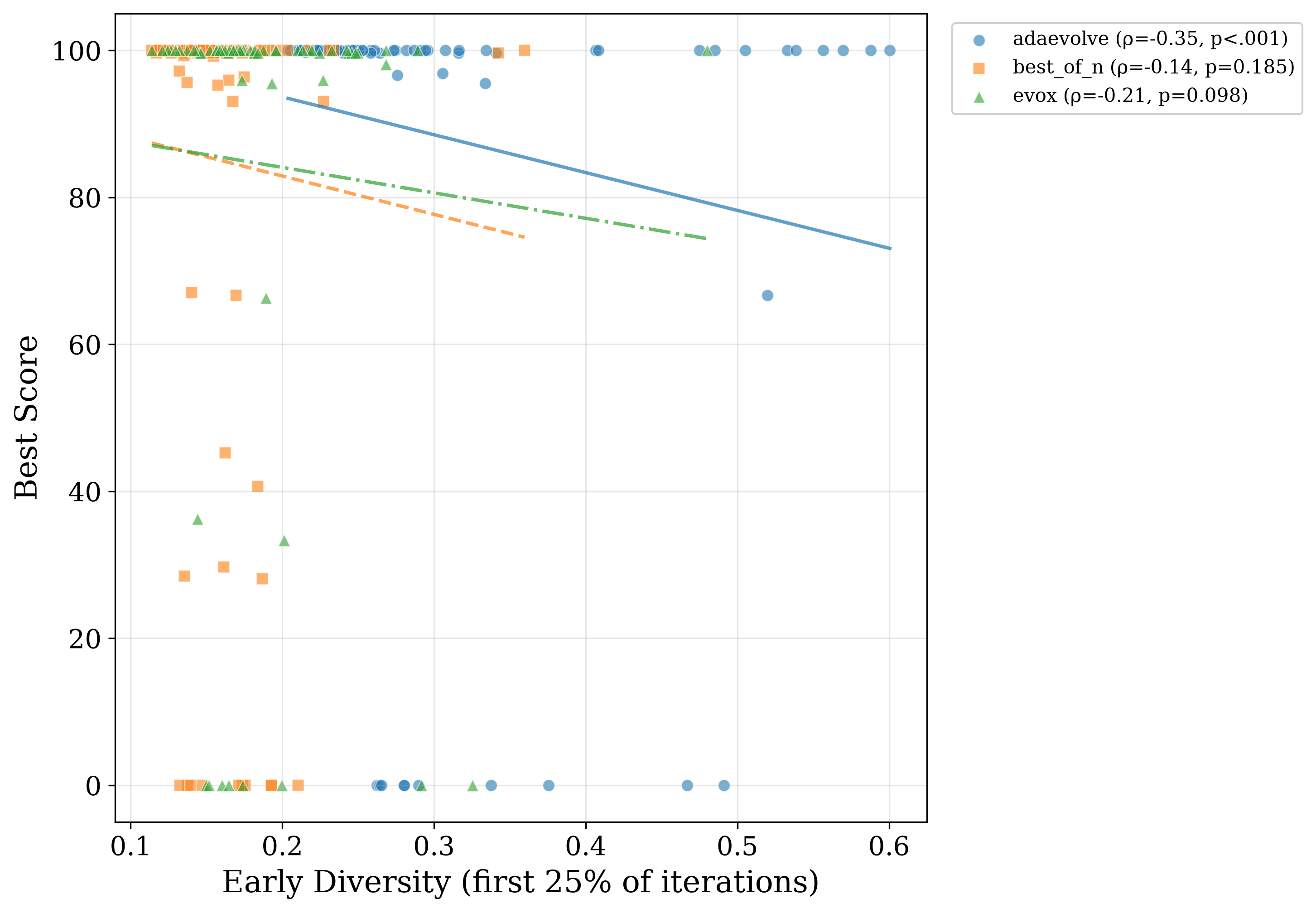}
\caption{Early diversity (first 25\% of iterations) vs.\ final score by mechanism for Bounded 2D Knapsack.}
\label{fig:rq3-knapsack-early-by-algorithm}
\end{figure}

\begin{figure}[h]
\centering
\includegraphics[width=0.7\linewidth]{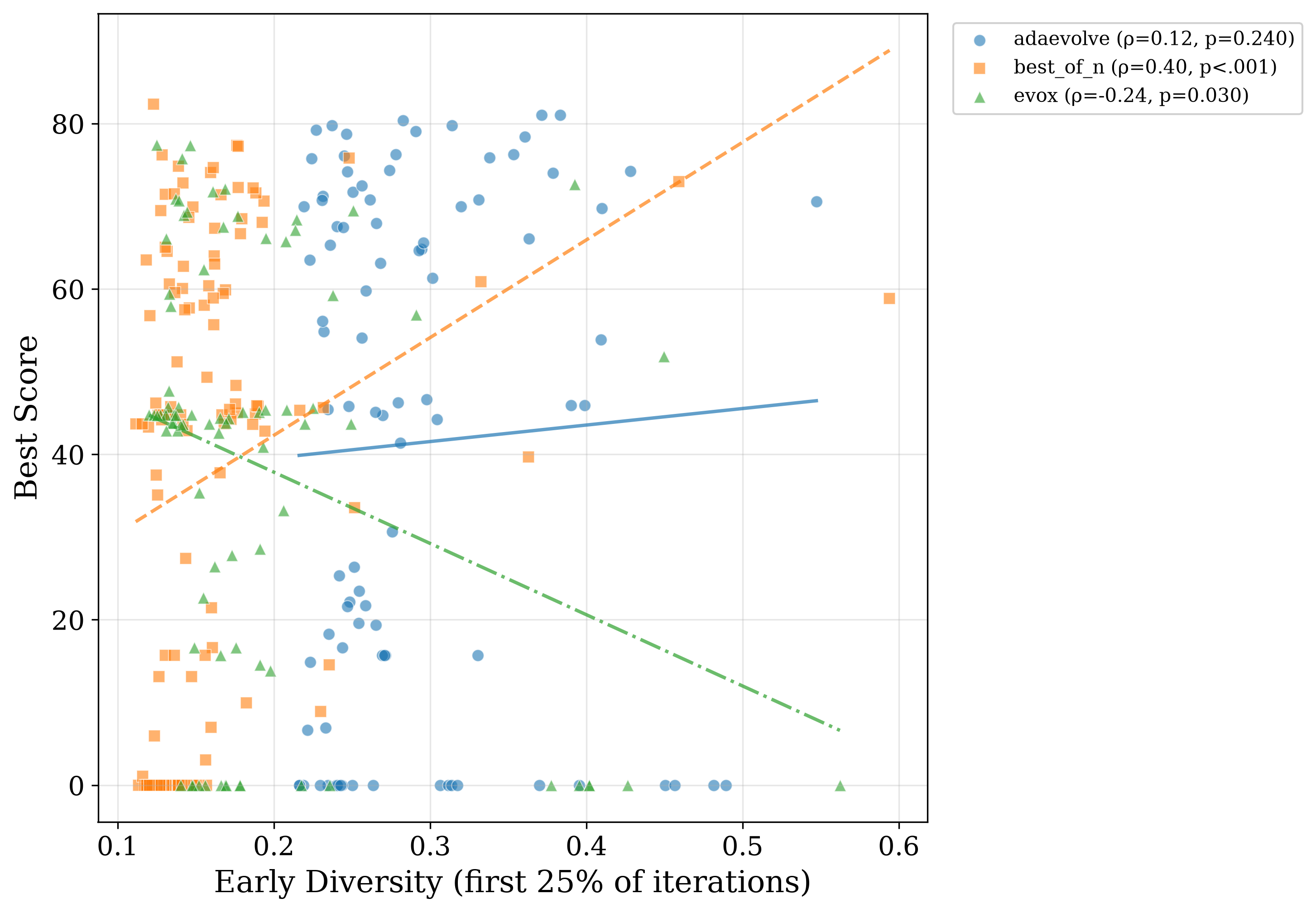}
\caption{Early diversity (first 25\% of iterations) vs.\ final score by mechanism for Polynomino Packing.}
\label{fig:rq3-early-by-algorithm}
\end{figure}

\begin{figure}[h]
\centering
\includegraphics[width=0.7\linewidth]{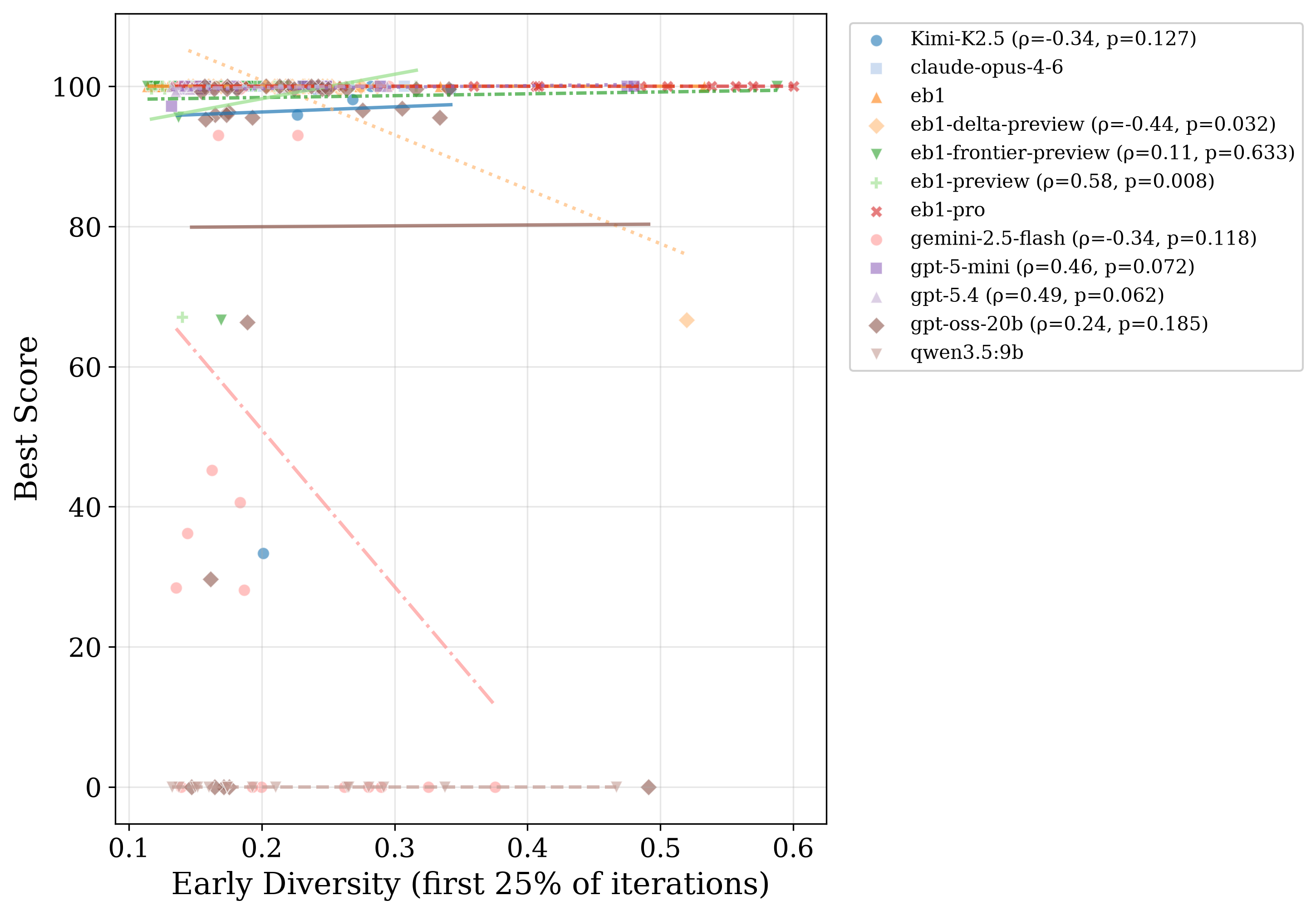}
\caption{Early diversity vs.\ final score by model for Bounded 2D Knapsack.}
\label{fig:rq3-knapsack-early-by-model}
\end{figure}

\begin{figure}[h]
\centering
\includegraphics[width=0.7\linewidth]{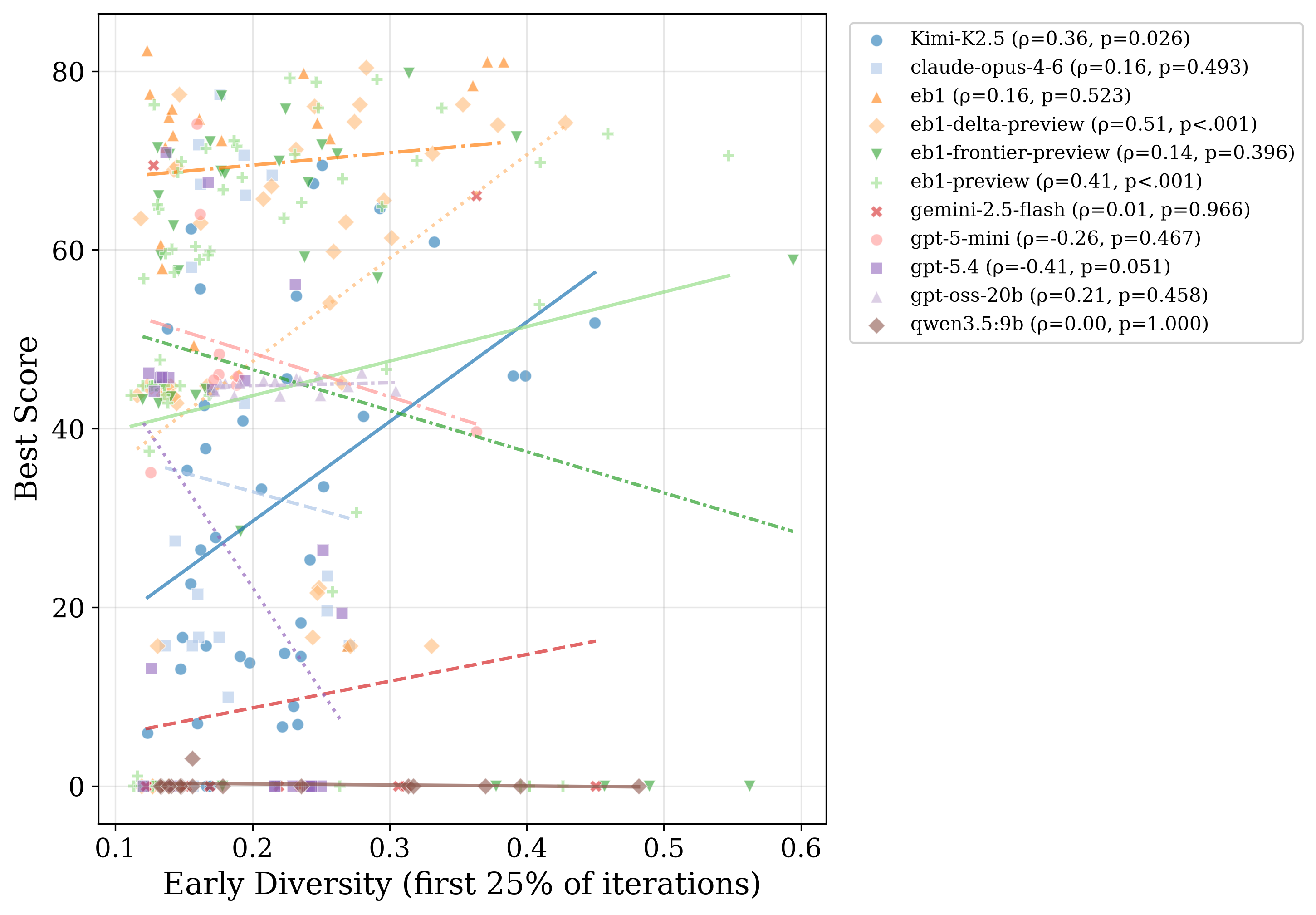}
\caption{Early diversity vs.\ final score by model for Polynomino Packing.}
\label{fig:rq3-early-by-model}
\end{figure}

Disaggregating by mechanism and model separately (Figures~\ref{fig:rq3-knapsack-early-by-algorithm}--\ref{fig:rq3-early-by-model}) shows the Polynomino positive direction holding for most subgroups, though some reverse the sign---notably EvoX shows a significant negative correlation ($\rho = -0.24$, $p = 0.030$) and gpt-5.4 trends negative ($\rho = -0.41$, $p = 0.051$). The pattern suggests that the predictive direction of early diversity may vary with configuration choice.

Early diversity thus serves as a context-dependent predictor of final outcome: it is positive on the multi-optima basin problem (favoring exploration) but carries no signal on the binary-basin problem, where the outcome is largely decided by whether a run reaches the correct basin rather than by how broadly it explores early on; certain configurations further reverse the problem-level direction. A complementary analysis of within-decile diversity across score tiers is presented in Appendix~\ref{sec:appendix-rq3-decile}.

\FloatBarrier

\subsection{Within-Decile Diversity}
\label{sec:appendix-rq3-decile}

We separately analyze diversity \emph{across} best candidates by partitioning runs into score deciles. Since our diversity metric operates on code embeddings, it captures similarity in algorithmic strategy and implementation rather than in the underlying mathematical solution. Neither problem shows a significant monotonic relationship between score decile and within-decile diversity: Bounded 2D Knapsack fluctuates without a clear trend (Spearman $\rho = -0.06$, $p = 0.87$), and Polynomino Packing is similarly flat, reaching its lowest diversity in the middle deciles rather than at the top (Spearman $\rho = 0.08$, $p = 0.83$). We therefore find no evidence that higher-scoring solutions converge to a narrower region of code space.

\begin{figure}[h]
\centering
\includegraphics[width=0.75\linewidth]{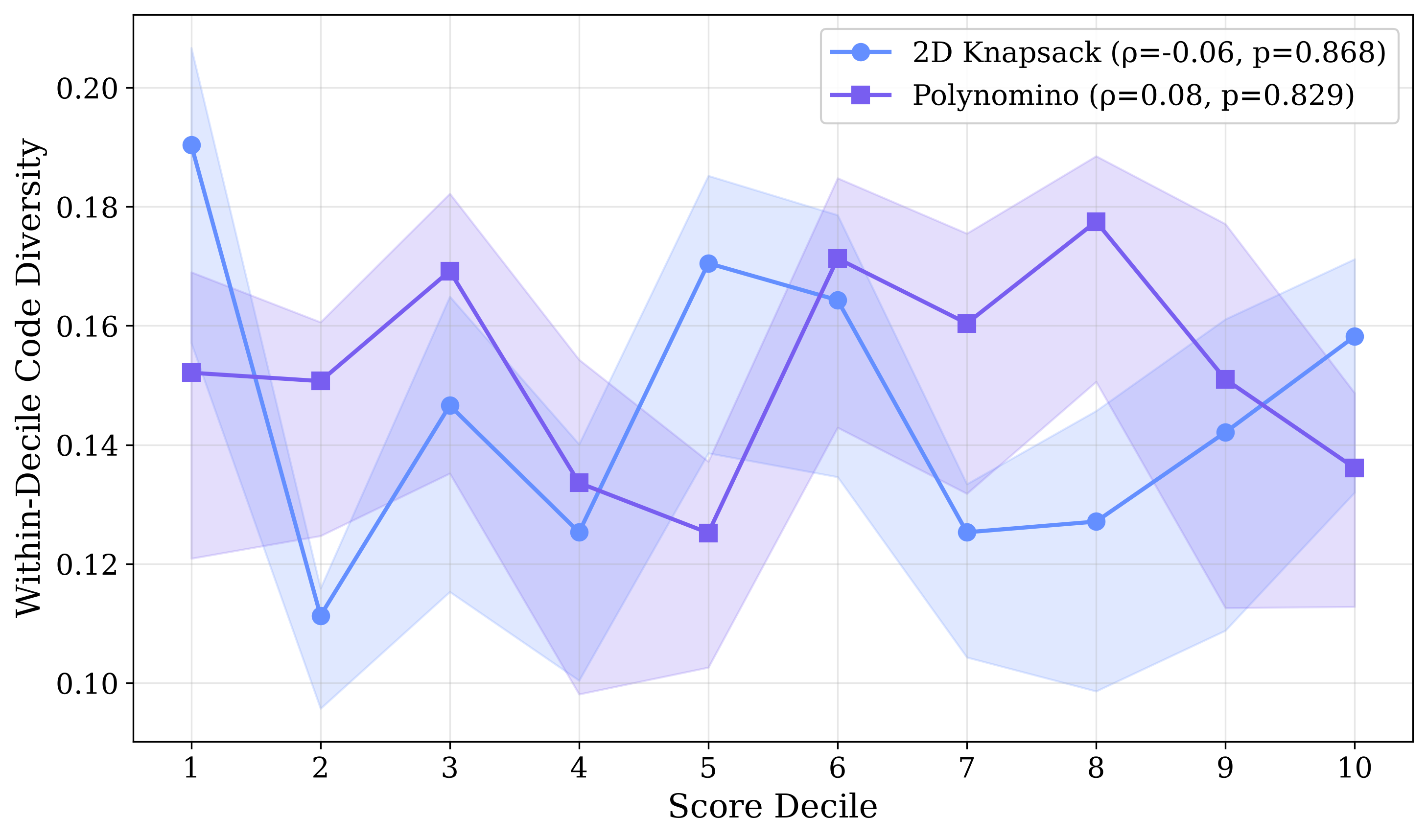}
\caption{Within-decile code diversity across score deciles. Runs are binned into ten equal-sized groups by best-candidate score (decile~1 = lowest, decile~10 = highest). Lower diversity indicates greater convergence among solutions in that performance tier.}
\label{fig:rq3-topk}
\end{figure}

\FloatBarrier

\section{Additional RQ3 Details: Context Composition}
\label{sec:appendix-rq4}

\subsection{Factor Definitions}
\label{sec:appendix-rq4-factors}

For each candidate $x$ with parent $p$ and context set $\text{ctx}(x)$, the four context-composition factors used in RQ3 are
\begin{align}
r^{+}(x) &= \frac{|\{c \in \text{ctx}(x) : \mathcal{A}(c) > \mathcal{A}(p)\}|}{|\text{ctx}(x)|}, \label{eq:better-ratio} \\
r^{-}(x) &= \frac{|\{c \in \text{ctx}(x) : \mathcal{A}(c) < \mathcal{A}(p)\}|}{|\text{ctx}(x)|}, \label{eq:worse-ratio} \\
\sigma^2_{\text{score}}(x) &= \text{Var}\bigl(\{\mathcal{A}(c) : c \in \text{ctx}(x)\}\bigr), \label{eq:score-div} \\
D_{\text{code}}(x) &= D\bigl(\text{ctx}(x)\bigr), \label{eq:code-div}
\end{align}
where $r^{+}$ and $r^{-}$ are the \emph{better} and \emph{worse context ratio}, $\sigma^2_{\text{score}}$ is the \emph{score diversity} in context, and $D_{\text{code}}$ is the \emph{code diversity} in context (Equation~\ref{eq:diversity}).

\subsection{Factor Importance by Model and Mechanism}

This subsection presents additional visualizations for the context-composition analysis of RQ3.

\begin{figure}[t]
\centering
\includegraphics[width=\linewidth]{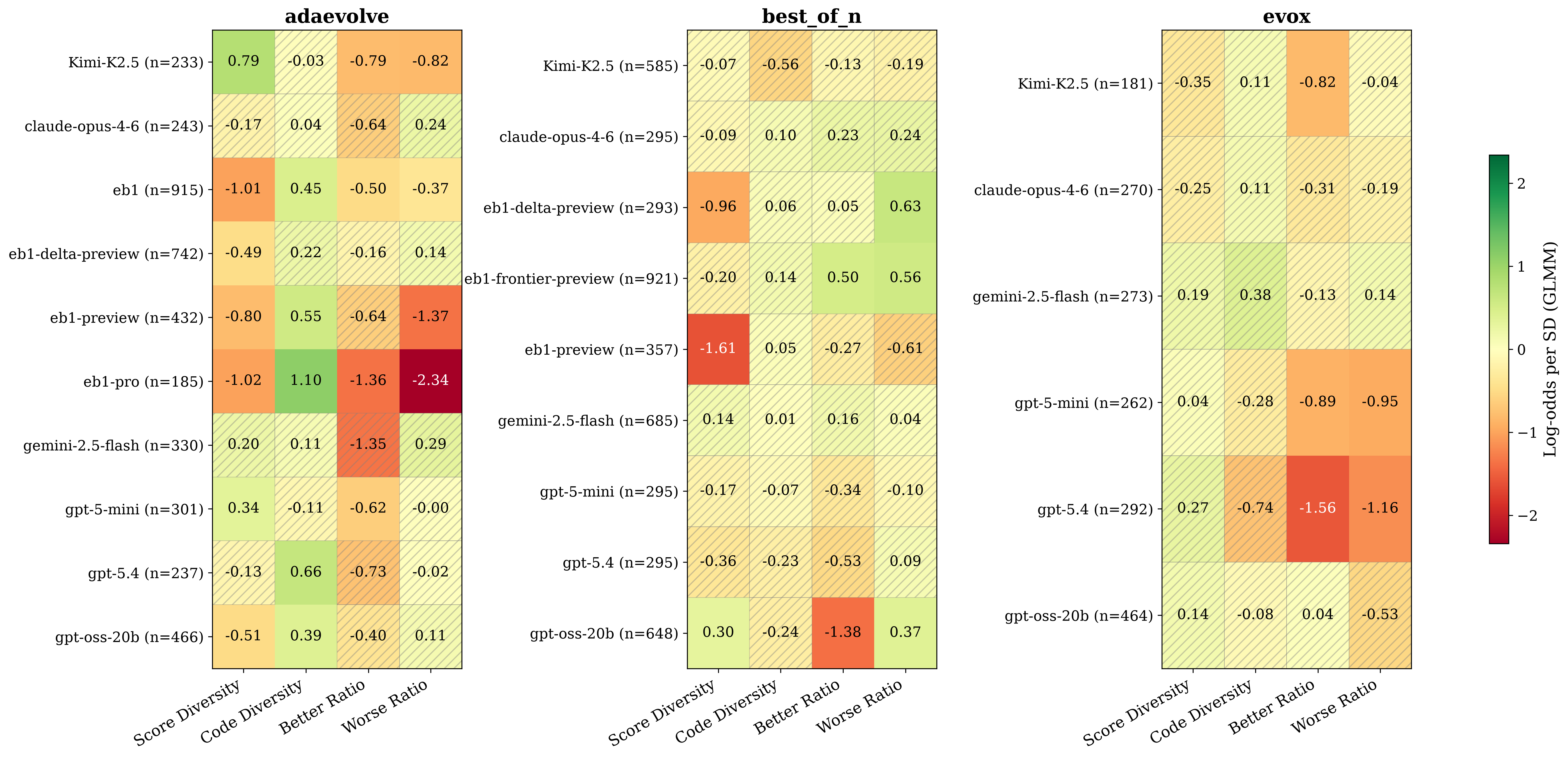}
\caption{Factor importance by model and mechanism for Bounded 2D Knapsack. Cells are GLMM log-odds per SD. Each panel shows one mechanism; rows are models, columns are factors. Hatched cells are not statistically significant ($p \geq .05$, Wald test). The negative direction is consistent across most model--mechanism combinations, with Gemini-2.5-Flash as a notable outlier.}
\label{fig:rq4-knapsack-by-model-algorithm}
\end{figure}

\begin{figure}[t]
\centering
\includegraphics[width=\linewidth]{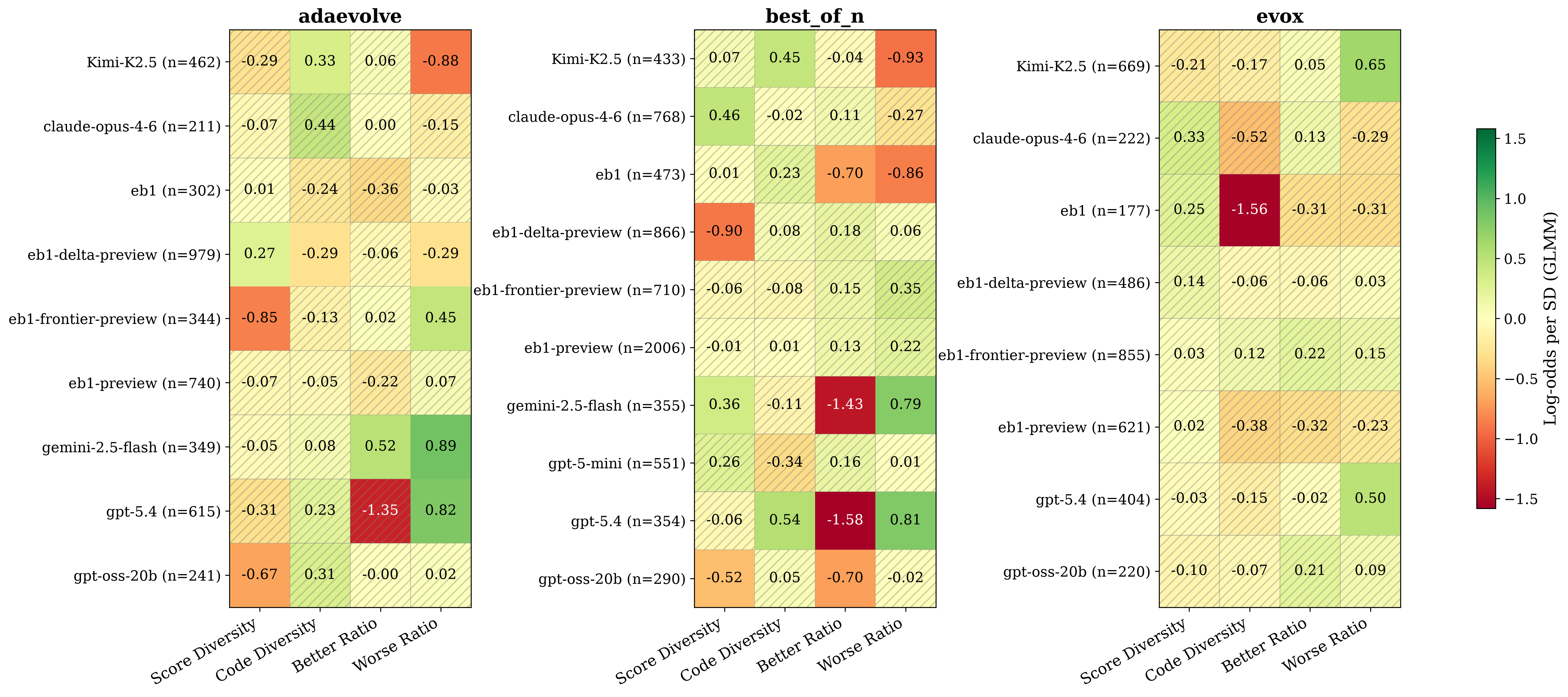}
\caption{Factor importance by model and mechanism for Polynomino Packing. Cells are GLMM log-odds per SD. Each panel shows one mechanism; rows are models, columns are factors. Hatched cells are not statistically significant ($p \geq .05$, Wald test). An overall more positive trend is observed.}
\label{fig:rq4-polynomino-by-model-algorithm}
\end{figure}

\FloatBarrier

\section{RQ4 Details}
\label{sec:appendix-rq5}

\begin{figure}[!ht]
\centering
\includegraphics[width=\linewidth]{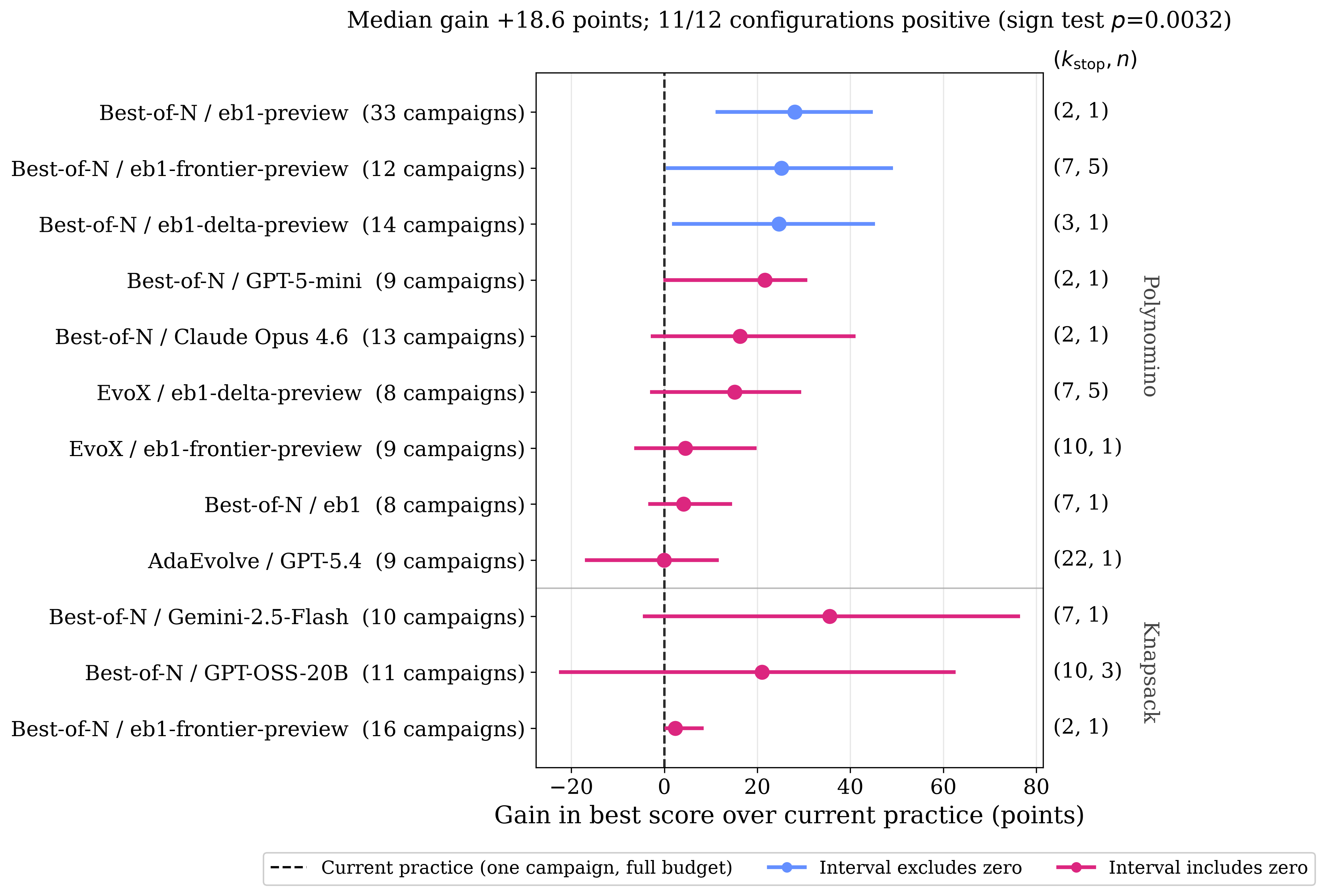}
\caption{Per-configuration gain of the ROAR budget policy over current practice (one campaign using the full budget) at an identical 50-iteration budget, for the 12 tested configurations. Points are leave-one-out estimates; bars are 95\% out-of-bag refit bootstrap intervals (1{,}000 draws), blue where the interval excludes zero. The right-hand column gives each configuration's fitted policy $(k_{\text{stop}}, n)$.}
\label{fig:rq5-forest}
\end{figure}

\subsection{Limitations of the Replay Evaluation}

\textbf{Replay from stored campaigns.} A relaunch is simulated by replaying another stored campaign of the same configuration. This treats campaigns of one configuration as interchangeable apart from their random seed. The policy is also fit on the same campaigns its relaunches are drawn from. Scoring on held-out campaigns keeps the comparison with current practice fair, but the size of the gain is likely optimistic. When the policy is fit on half of a configuration's campaigns and relaunches are drawn from the other half, the gain is smaller but remains positive in 10 of 12 configurations.

\textbf{Tested configurations.} We test configurations with at least 8 campaigns of at least 50 iterations and a best-score standard deviation of $\sigma_s \geq 2$ points on the 0--100 scale. The cutoff is operational rather than tuned. Of the 17 configurations with enough campaigns, the five excluded have $\sigma_s \leq 1.3$, four of them with every campaign at or within a point of the maximum score of 100, while every tested configuration has $\sigma_s \geq 8$, so any cutoff between those values selects the same twelve.

\textbf{Corpus size.} Twelve configurations, three of them on Knapsack, are enough to show that the gain is consistently positive but not to pin down its size, as the wide Knapsack interval reflects. The configurations themselves hold 8--33 campaigns. Percentile bootstrap intervals at these sizes are known to under-cover, and each refit scores an out-of-bag set of only a few campaigns, so the per-configuration bars in Figure~\ref{fig:rq5-forest} are anti-conservative and the blue coloring should be taken as suggestive rather than as a strict test. The claim we rely on is the direction of the gain, positive in 11 of 12 configurations, together with the problem-level intervals, which pool 115 and 37 campaigns. Each campaign contributed to ROAR tightens these estimates at no added cost.

\FloatBarrier

\end{document}